\documentclass[runningheads]{llncs}

\usepackage{eccv}

\usepackage{eccvabbrv}

\usepackage{graphicx}
\usepackage{booktabs}
\usepackage{multirow}
\usepackage{wrapfig}

\usepackage[accsupp]{axessibility}  

\usepackage[pagebackref,breaklinks,colorlinks,citecolor=eccvblue]{hyperref}
\usepackage{hyperref}

\usepackage{orcidlink}

\begin{document}

\title{Semantic Compositions Enhance Vision-Language Contrastive Learning} 

\titlerunning{CLIP-$\mathcal{C}$: CLIP Compositions}

\author{Maxwell Mbabilla Aladago\inst{1} \and
Lorenzo Torresani\inst{1,2} \and
Soroush Vosoughi\inst{1}}

\authorrunning{M.~Aladago et al.}

\institute{Dartmouth College, Hanover NH 03755, USA \and Meta, FAIR\\
\email{\{maxwell.m.aladago.gr, LT, soroush.voshoughi\}@dartmouth.edu}}

\maketitle


\begin{abstract}
  In the field of vision-language contrastive learning, models such as CLIP capitalize on matched image-caption pairs as positive examples and leverage within-batch non-matching pairs as negatives. This approach has led to remarkable outcomes in zero-shot image classification, cross-modal retrieval, and linear evaluation tasks. We show that the zero-shot classification and retrieval capabilities of CLIP-like models can be improved significantly through the introduction of semantically composite examples during pretraining. Inspired by CutMix in vision categorization, we create semantically composite image-caption pairs by merging elements from two distinct instances in the dataset via a novel procedure. Our method fuses the captions and blends 50\% of each image to form a new composite sample. This simple technique (termed CLIP-$\mathcal{C}$ for CLIP Compositions), devoid of any additional computational overhead or increase in model parameters, significantly improves zero-shot image classification and cross-modal retrieval. The benefits of CLIP-$\mathcal{C}$ are particularly pronounced in settings with relatively limited pretraining data.
\end{abstract}

\section{Introduction}
\label{sec:intro}

Recent advancements in vision-language pretraining have propelled a multitude of tasks, including zero-shot image classification~\cite{radford2021learning, JiaAlign2021, declip}, video understanding~\cite{xu-etal-2021-videoclip, zhao2022lavila}, and various multi-modal applications~\cite{MaMMUT, yu2022coca, naeem2023silc}. These successes echo the transformative trajectory initiated by large-scale pretraining efforts in Computer Vision (CV)~\cite{alexnet, resnet} and later in Natural Language Processing (NLP)~\cite{bert, gpt1, radford2019language, gpt3}. A prominent recent example in the vision-language interplay is the Contrastive Language-Image Pre-training (CLIP) model~\cite{radford2021learning}, which has become a benchmark for language-supervised training.

The objective of contrastive language-supervised pretraining is straightforward yet powerful: to align embeddings of corresponding image-text pairs in a shared embedding space while distancing the non-matching pairs~\cite{simclr, SimSiam2021, moco}. CLIP has pioneered this direction with a dual-encoder framework, training on an expansive dataset of image-caption pairs sourced from the internet, using a bidirectional contrastive loss~\cite{oord2019representation}.

Subsequent studies aimed at improving the data efficiency of CLIP have introduced supplementary objectives, such as within-modality self-supervision~\cite{mu2021slip, declip}, multi-crop supervision~\cite{declip}, and the use of captions generated by large language models~\cite{fan2023improving, lai2023scarcity, zhao2022lavila}. These methods typically necessitate additional computational steps, such as multiple forward passes or the inclusion of extra encoders.

\begin{figure*}[t]
    \centering
    \includegraphics[width=0.9\textwidth]{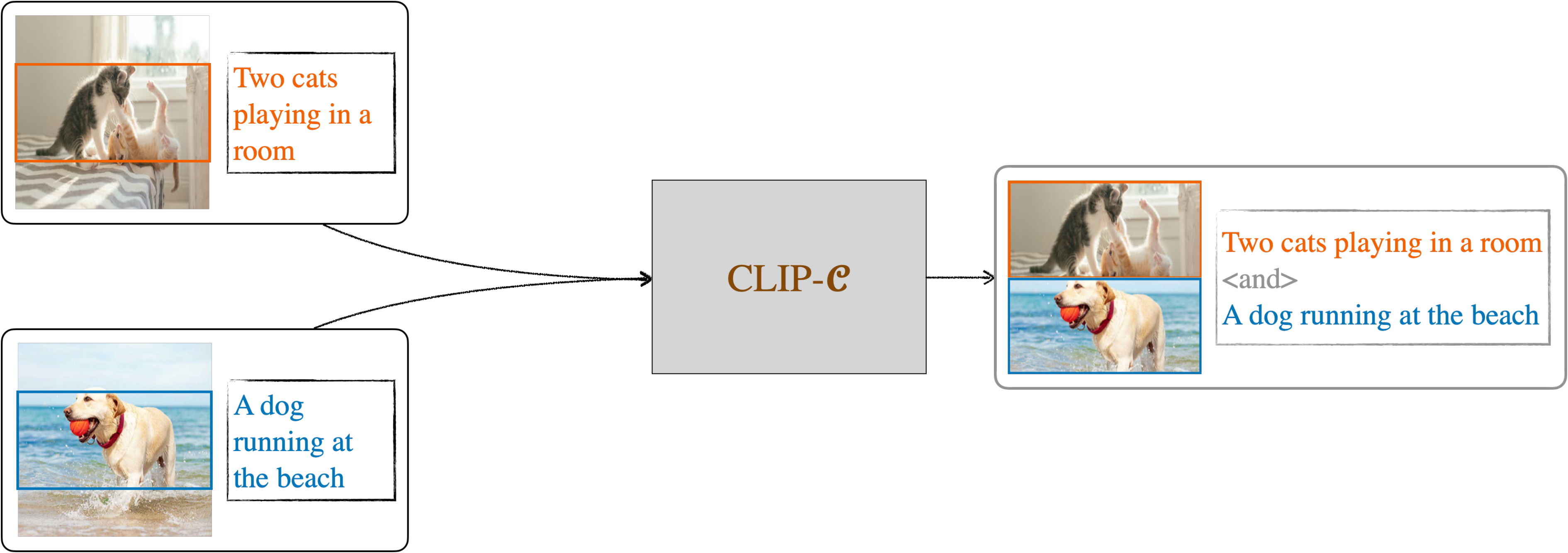}
    \caption{\textbf{CLIP-$\mathcal{C}$}:  We use the center half crops spanning the width (as in this illustration) or the height of the image. The captions are concatenated with the delimiter ``and''. We vary the positions of the captions on either side of the conjunction, \ie, the output caption can be either (a) $\{\text{caption1 } \text{and } \text{caption2}\}$ or (b) $\{\text{caption2 } \text{and } \text{caption1}\}$. We emphasize that only a fraction of the batch in each iteration constitute composite samples. The colored boxes and texts shown here are for illustrative purposes.}
    \label{fig:main-method}
\end{figure*}

Our contribution, termed CLIP-$\mathcal{C}$ (illustrated in \cref{fig:main-method}), enhances data efficiency through an innovative but straightforward compositional approach, merging original image-caption pairs into novel compound examples. This approach builds upon the achievements of CutMix~\cite{yun2019cutmix} in the domain of vision categorization tasks, adapting it to the vision-language pretraining context. This adaptation results in significant enhancements in model performance across various downstream tasks, achieved without incurring extra computational costs or increasing model complexity.

%

%


Our technique involves composing together two image-caption pairs from the dataset. This is done by conjoining the captions with ``and'' serving as the conjunction and merging the central crops of both images. The result is a set of compound instances that embody a broader array of concepts than the individual pairs, presenting the model with expanded semantic challenges that drive learning (\cref{fig:per-epoch-zero-shot-acc}). The two image-caption pairs constituting the composite instance are sampled dynamically in each iteration based on a predefined probability, empowering the model to uncover novel combinations of examples throughout training.

Distinguishing itself from stylistic variation methods that manipulate single examples, CLIP-$\mathcal{C}$ leverages ``semantic composition'' to introduce contextually varied training examples. Interestingly, we discover that the benefits of CLIP-$\mathcal{C}$ arise from other factors besides the diverse nature of these novel semantic associations. Surprisingly and very counter-intuitively, we find that the model learns to match compound image-caption pairs more easily than the original plain image-caption pairs. This then initiates a positive spillover effect where the model is able to learn better representations of plain unmodified examples in CLIP-$\mathcal{C}$ compared to CLIP.  






Like~\cite{fan2023improving}, our approach improves the performance of CLIP in poor data scenarios. However, we enhance the data with compositions of both the images and captions without any reliance on external systems. As a result, we can efficiently and flexibly generate novel captions and images online during training. 
Moreover, our method does not increase the batch size or the number of iterations needed, maintaining operational parity with CLIP. Indeed, we demonstrate that training CLIP longer or with a higher batch size than CLIP-$\mathcal{C}$ is not sufficient to close the performance gap between the two methods. 

In downstream applications, CLIP-$\mathcal{C}$ exhibits a competitive edge, surpassing CLIP by over 5\% in cross-modal retrieval accuracy on Flickr30k~\cite{flickr30} and showing substantial improvements on MS-COCO~\cite{ms-coco} in both image-to-text and text-to-image retrieval tasks. Additionally, our model demonstrates impressive gains in zero-shot classification, with a 2\% increase on ImageNet~\cite{imagenet}, and superior linear evaluation results without necessitating any additional model parameters, memory, or dependence on external language processing systems. Even when evaluated on relatively large datasets such as CC12M~\cite{cc12m} and RedCaps~\cite{desai2021redcaps}, CLIP-$\mathcal{C}$ still outperform the baseline CLIP in cross-modal retrieval and zero-shot classification tasks, albeit with decreased margins. 


Finally, we believe that CLIP-$\mathcal{C}$ will be particularly beneficial in contexts where image-text datasets do not exist in large quantities or are not easily accessible (\eg, medical images, satellite images, etc.). However, it is not feasible to carry out comprehensive evaluations in these domains precisely because there are no established benchmarks of images with captions for them. Thus, we use evaluations on medium-size Web-derived datasets to demonstrate the potential value of our approach in application scenarios where in-domain data is not as abundantly available as for general natural images. 






\section{Related Works}
\label{sec:related-works}
The use of language as an effective supervisory signal for learning visual representations has a rich history in machine learning~\cite{Quattoni2007, devise2013, Joulin2016, Li2017, radford2021learning}. Early influential works such as DeViSE~\cite{devise2013} first learned semantic relations using unannotated textual data before mapping the images into that semantic space using class labels.
More recently, models like CLIP~\cite{radford2021learning}, ALIGN~\cite{JiaAlign2021}, and others~\cite{mu2021slip, declip, xu2023metaclip, openclip} further improved the capabilities of joint vision-language embedding models by training on massive image-text paired datasets contrastively using the InfoNCE loss~\cite{oord2019representation}. Our work aligns with these prior arts but focuses on incorporating semantic compositions during pretraining to improve data efficiency and enhance performance. 

Both CLIP~\cite{radford2021learning} and ALIGN~\cite{JiaAlign2021} use huge datasets ---400 million and 1B image-text pairs for CLIP and ALIGN, respectively. DeCLIP~\cite{declip} improves the data efficiency of CLIP by incorporating several training objectives, including self-supervision within each modality~\cite{SimSiam2021, bert}, nearest-neighbor supervision, and multi-view supervision~\cite{multiview}. SLIP~\cite{mu2021slip}, on the other hand, adds image self-supervision, SimCLR~\cite{simclr}, to the language supervision. In~\cite{Otter}, Wu~\etal show good zero-shot results in the low data regime through soft image-text matches via optimal transport distillation. These methods, however, require multiple passes through the image encoder~\cite{mu2021slip, declip} for each update or a first-in-first-out feature queue~\cite{declip} to generate the representations for the extra objectives. Our method is free of these additional complexities.




 Most similar to our work are data augmentation methods such as CutMix~\cite{yun2019cutmix} and MixUP~\cite{zhang2018mixup} which have been very effective in training categorization models in computer vision.  Our work brings the benefits of these established augmentation techniques in image understanding to the vision-language joint-embedding space. In addition to the incorporation of language, our method differs from CutMix by concatenating the image crops instead of pasting one crop on the other. Additionally, we train our models using contrastive loss. 
As our method is a pre-training mechanism, we do not discuss works~\cite{jiang2023comclip, ITMix, lit2022, li2023blip2, yuksekgonul2023when} that use open-source CLIP checkpoints transfer learning scenarios. 


\section{Method}
\label{sec:method-main}

This section covers a background of the baseline method as well as the core components of CLIP-$\mathcal{C}$'s framework. 
\subsection{Background}
Contrastive Language-Image Pre-training (CLIP) from Radford~\etal\cite{radford2021learning} has emerged as a highly successful approach for training vision-language models. CLIP is a dual encoder model with separate encoders $f_I$ and $f_T$ for extracting visual and textual features respectively. It also has two dedicated projection functions $g_I$ and $g_T$ that map the outputs of the encoders to a shared embedding space. Given a batch of $B$ 
images and text pairs $\left\{x_I^{(i)}, x_T^{(i)}\right\}_{i=1}^B$ in each training step, CLIP  computes the embeddings  $z_I^{(i)} = g_I\left(f_I\left(x_I^{(i)}\right)\right)$ and $z_T^{(i)} = g_T\left(f_T\left(x_T^{(i)}\right)\right)$ where $z_I^{(i)} \in \mathbb{R}^d$ represents the normalized features of image $x_I^{(i)}$.  $z_T^{(i)} \in \mathbb{R}^d$ denotes the normalized features  of the corresponding caption $x_T^{(i)}$. The loss is evaluated using InfoNCE~\cite{oord2019representation} whereby matching image-text pairs 
$\{x_{I}^{(i)}, x_{T}^{(i)}\}$ constitute the positive samples and non-matching pairs $\{x_{I}^{(i)}, x_{T}^{(j)}\}\quad \forall j \neq i$ form the negative examples. A bidirectional loss is computed as 
\begin{align}\label{image-to-text-loss}
    \mathcal{L}_{I_2T} &= -\frac{1}{B}\sum_{i = 1}^B\log\frac{\exp\left(\frac{1}{\tau}\text{sim}\left(z_I^{(i)}, z_T^{(i)}\right)\right)}{\sum_{j=1}^B\exp\left(\frac{1}{\tau}\text{sim}\left(z_I^{(i)}, z_T^{(j)}\right)\right)}
\end{align}
\begin{align}\label{text-to-image-loss}
    \mathcal{L}_{T_2I} &= -\frac{1}{B}\sum_{i = 1}^B\log\frac{\exp\left(\frac{1}{\tau}\text{sim}\left(z_I^{(i)}, z_T^{(i)}\right)\right)}{\sum_{k=1}^B\exp\left(\frac{1}{\tau}\text{sim}\left(z_I^{(k)}, z_T^{(i)}\right)\right)}
\end{align}
where temperature $\tau$ is typically a learnable parameter used to scale the logits. $\tau$ is fixed in all of our ablation experiments as it has a noticeable impact on the model~\cite{mindgap} which makes comparisons across different experiments difficult. $\text{sim}(\cdot, \cdot)$ is a similarity function measuring the distance between the features. In CLIP~\cite{radford2021learning} and our experiments, $\text{sim}(\cdot, \cdot)$ is set as the dot product function. 
The total loss is an average of the two losses in \cref{image-to-text-loss} and \cref{text-to-image-loss}: 
\begin{align}\label{final-loss}
    \mathcal{L} = (\mathcal{L}_{I_2T} + \mathcal{L}_{T_2I})/2 \ .  
\end{align}
\subsection{CLIP-\texorpdfstring{$\mathcal{C}$}{$C$}}
\label{sec:method-ours}
In each training step, CLIP-$\mathcal{C}$ samples a batch of examples of size $B$, $\left\{\hat{x}_I^{(i)}, \hat{x}_T^{(i)}\right\}_{i=1}^B$. Any given paired instance $\left(\hat{x}_I^{(i)}, \hat{x}_T^{(i)}\right)$ is either the original example  $\left({x}_I^{(i)}, {x}_T^{(i)}\right)$ or a composition of that example and another example $\left({x}_I^{(i^\prime)}, {x}_T^{(i^\prime)}\right), i \neq i^\prime$, drawn from the dataset. Note that index $i^\prime$ is taken with respect to the dataset size and not the batch size $B$, \ie, sample $i^\prime$ may not be present in the current mini-batch. The proportion of composed samples in any mini-batch is controlled by a sampling rate hyper-parameter $\rho$. The impact of this parameter is discussed in \cref{subsec:composition-ratio}. 

In the case whereby $\left(\hat{x}_I^{(i)}, \hat{x}_T^{(i)}\right)$ is a composite sample, the new caption $\hat{x}_T^{(i)}$ is a concatenation of the two original captions involved: $\hat{x}_T^{(i)} = [x_T^{(i)}, x_T^{(i^\prime)}]$ where $[\cdot, \cdot]$ is a string concatenation function with the word ``and'' as a  conjunction. The positions of the captions on either side of this conjunction change, with $x_T^{(i)}$ appearing first fifty percent of the time.

The new image is composed of the center half crops spanning either the height or the width of each image. For example, if the images have resolution $(S \times S)$, either $(\frac{S}{2} \times S)$ or $(S \times \frac{S}{2})$  center crops are taken from both images and concatenated as illustrated in \cref{fig:main-method}. We experiment with other forms of image augmentation methods such as MixUP\cite{zhang2018mixup} and CutMix\cite{yun2019cutmix} in \cref{tab:zero-shot-image-mixup}. 

After assembling the mini-batch as described above, CLIP-$\mathcal{C}$ proceeds to extract the image and text features as in CLIP: $\hat{z}_I^{(i)} = g_I\left(f_I\left(\hat{x}_I^{(i)}\right)\right)$ and $\hat{z}_T^{(i)} = g_T\left(f_T\left(\hat{x}_T^{(i)}\right)\right)$. With $\hat{z}_I^{(i)}$ and $\hat{z}_T^{(i)}$ computed, \cref{image-to-text-loss}, \cref{text-to-image-loss}, and \cref{final-loss} are used to compute the InfoNCE loss. 

The sampling strategy CLIP-$\mathcal{C}$ employs exposes the model to a much higher diversity of images and their corresponding captions compared to the vanilla pretraining pipeline. As a result, we observe much more significant improvements in downstream transfer when the pretraining dataset is small. It is reasonably expected that relatively larger datasets such as RedCaps~\cite{desai2021redcaps} are already sufficiently diverse and, therefore, may not benefit from our method. Nonetheless, CLIP-$\mathcal{C}$ still does better than CLIP on these large datasets.

\section{Experimental Setup}
\label{sec: experimental-setup-main}
All our experiments use the CLIP framework due to its demonstrated effectiveness, simplicity, and widespread usage. We emphasize that we do not use pretrained CLIP checkpoints from prior works as our method is a pretraining mechanism. Thus, we retrain CLIP on our pretraining datasets and compare it to our approach. Finally, due to resource constraints, we conduct our experiments in the low data and small model regimes. Consequently, we are unable to compare with prior large-scale training systems.

\textbf{Pretraining Datasets. }We use three widely adopted web-crawled datasets of varying sizes and distributions for our experiments: Conceptual Captions~\cite{cc3m}, Conceptual 12M~\cite{cc12m}, and RedCaps~\cite{desai2021redcaps}. These three datasets together enable us to assess the effectiveness of our method across pretraining datasets of different sizes and qualities.

\textbf{Models. } We use Vision Transformer~\cite{dosovitskiy2021an} models of various sizes as in~\cite{mu2021slip}. The vision encoder is set to ViT-S/16~\cite{deit} in all our ablation experiments unless explicitly specified otherwise. We use ViT-B/16~\cite{dosovitskiy2021an, deit} as the image encoder to demonstrate the efficacy of our method at scale as we are unable to run much bigger models such as ViT-L/16 because of resource constraints. The text encoder in all our experiments is set to the 38M parameter text Transformer model from~\cite{radford2021learning}. Following previous methods, Byte-Pair encoding is used for tokenization with a context length of 77 and a vocabulary size of 49k. Finally, we fixed the temperature parameter at $0.01$, the maximum value used in CLIP~\cite{radford2021learning}.

\textbf{Hyper-parameters.} We train all our models using PyTorch~\cite{pytorch} with a global batch size of $2,048$ split across 8 GPUS in a single machine. AdamW~\cite{adamw} is the optimizer during pretraining.  All models are pretrained for 40 epochs using a cosine decay learning rate schedule with a base rate of $0.003$, a warm-up period of $5$ epochs, and a final learning rate of $1e^{-5}$. The weight decay parameter is always set to $0.1$. Random cropping is the only augmentation applied to the images during pretraining. We refer the reader to the Supplemental for more detailed information about these and other hyper-parameters.

\textbf{Evaluation}. We perform zero-shot evaluation on several classification benchmarks using class names and prompts provided by~\cite{radford2021learning, mu2021slip}. First, the embeddings for all classes in a given benchmark are computed with each class embedding being an ensemble of multiple prompt templates. The highest cosine similarity between the image embedding and the class embeddings is then used as the zero-shot prediction. 

We test our model on eleven downstream datasets including ImageNet~\cite{imagenet}, CIFAR-10~\cite{cifar}, CIFAR-100~\cite{cifar}, Caltech-101~\cite{caltech101}, Oxford Pets~\cite{oxfordpets}, Country211~\cite{radford2021learning}, DTD~\cite{dtd}, Sun397~\cite{sun397}, STL-10~\cite{coates2011stl10}, RESISC-45~\cite{resisc45}, and EuroSAT~\cite{helber2017eurosat}. Following previous works~\cite{mu2021slip, fan2023improving}, we use ``mean per class accuracy''  as the metric for Oxford Pets and Caltech-101. Accuracy is the metric for all other datasets. In addition to the zero-shot analysis, we also conduct zero-shot retrieval in \cref{sec:retrieval}, and linear probing evaluations in \cref{sec:linear-probe}. For the linear probing experiments, we use the standard ``train'' and ``test'' splits for training and evaluation whenever possible. In instances where the standard splits are not present (RESISC-45~\cite{resisc45}, and EuroSAT~\cite{helber2017eurosat}), we randomly split the dataset into an 80\%-20\% ratio for training and testing respectively.


 
\section{Results}
This section outlines our key comparisons between CLIP and CLIP-$\mathcal{C}$ (our method) on zero-shot image classification, cross-modal retrieval, and linear probing. However, we explain first why our method works. 

 \begin{figure}[t]
    \begin{minipage}{0.47\textwidth}
     \centering
     \includegraphics[width=1.0\linewidth]{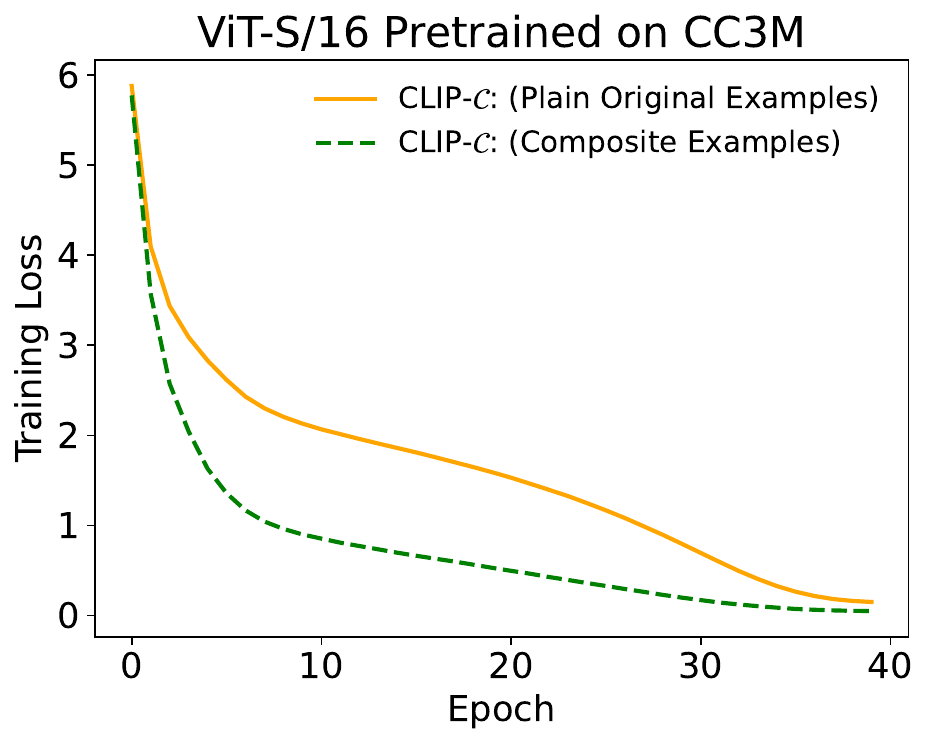}
     \caption{Counter-intuitively, the model learns to match the composite examples faster compared to the plain instances. }
     \label{fig:per-epoch-losses}
    \end{minipage}\hfill
    \begin{minipage}{0.47\textwidth}
     \centering
     \includegraphics[width=1.0\linewidth]{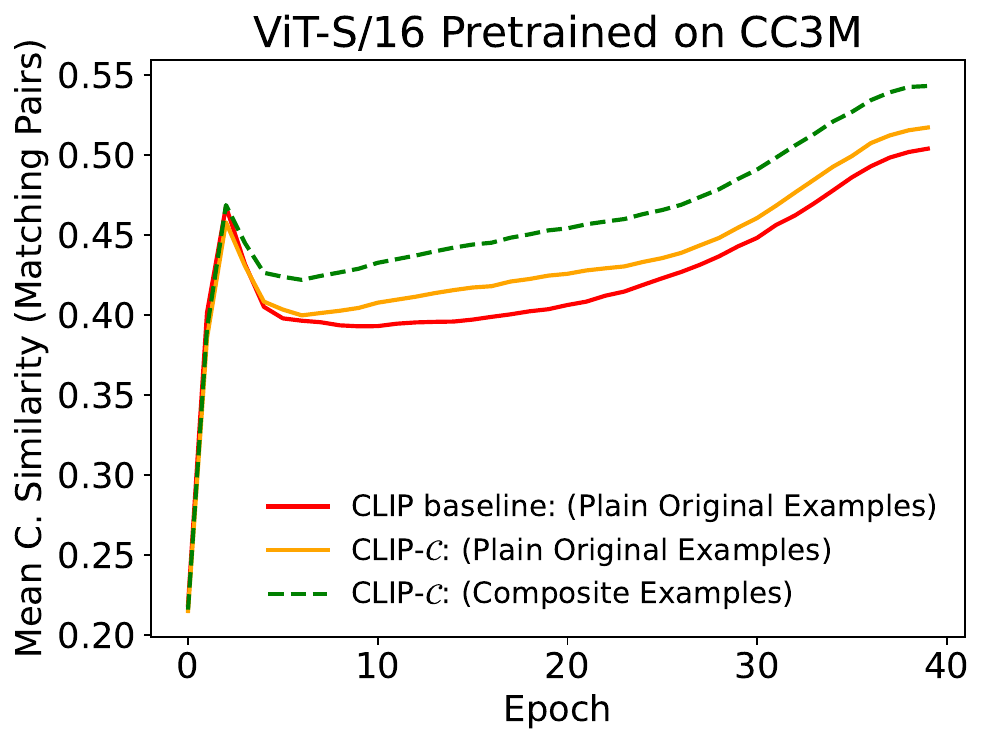}
     \caption{CLIP-$\mathcal{C}$ generally produces higher cosine similarity for matching pairs than CLIP.}
     \label{fig:per-epoch-cs}
    \end{minipage}
 \end{figure}
 \subsection{Why is CLIP-\texorpdfstring{$\mathcal{C}$}{$C$} an Effective Method?}
 \label{sec:effectiveness}
 Why will combining multiple different image-caption pairs into single instances during pretraining lead to improvements in downstream evaluations?  In other words, why will CLIP-$\mathcal{C}$ work? To investigate this salient question, we examined the pretraining losses and cosine similarities of both the composite examples and plain examples as the model evolves. This fine-grained tracking of training mechanics provides insights into how the model handles plain simple examples versus composite examples, and whether there are any differences between the two groups.   
 
 Contrary to expectation that compound examples will be the more challenging to the model (since they are multiple examples condensed into single instances), we observed precisely the opposite: as shown in~\cref{fig:per-epoch-losses}, the loss on the composite examples is lower than the loss on plain examples especially in the early stages. Our hypothesis for this empirical observation is that the model more easily recognizes compound image-caption pairs because they tend to be structurally different from plain examples. The more interesting development arising from this phenomenon, however, is that the model is encouraged to dedicate more effort into learning the plain examples in CLIP-$\mathcal{C}$ compared to CLIP as seen in~\cref{fig:per-epoch-cs}. We believe this elevated learning of plain examples together with the use of dynamic semantic compositions (See~\cref{sec:ablations-main}) all contribute to the superior capabilities of our method as discussed in the next sections. 
 
 

\begin{table*}[t]
    \caption{\textbf{Zero-shot Image Classification}: CLIP-$\mathcal{C}$ is our method. CLIP is a the model from~\cite{radford2021learning} trained in our setting. CC3M CLIP-$\mathcal{C}$ models use $\rho = 0.3$ while CC12M and RedCaps models use $\rho = 0.15$.  Bold numbers are the best in each dataset and architecture comparison. }
     \label{tab:zero-shot-results-main}
     \centering
     \begin{tabular}{l|l|ccccccccccc|c}
     \rotatebox{90}{PT Dataset} & \rotatebox{90}{Method} & \rotatebox{90}{Food-101} & \rotatebox{90}{CIFAR-10} & \rotatebox{90}{CIFAR-100} & \rotatebox{90}{Caltech-101} & \rotatebox{90}{Pets} &\rotatebox{90}{DTD} & \rotatebox{90}{Country211} & \rotatebox{90}{Sun397}& \rotatebox{90}{STL-10} &  \rotatebox{90}{RESISC45} & \rotatebox{90}{EuroSAT} & \rotatebox{90}{ImageNet}\\
     \toprule
     \multicolumn{14}{c}{\textit{Vision Encoder: ViT-S/16}} \\
    \midrule
    \multirow{2}{*}{CC3M} & CLIP & 11.6 &  56.1 & 22.7   &  46.9 & 12.9 & 10.5  & 0.6 & 20.5 & 77.0 & 24.5 & 23.7 & 18.5\\
    & CLIP-$\mathcal{C}$ & \bf{15.1} & \bf{66.4} &  \bf{26.9} & \bf{51.9}  & \bf{14.5} & \bf{14.8} & \bf{0.7} & \bf{27.2} & \bf{84.6}  & \bf{25.4} & \bf{30.7} &  \bf{20.5}\\
    \midrule
    \multicolumn{14}{c}{\textit{Vision Encoder: ViT-B/16}} \\
     \midrule
     \multirow{2}{*}{CC3M} & CLIP & 13.8 & 54.8 &  20.4 &  49.8 & \bf{14.9} & 12.2 & 0.7 & 21.9 &  76.0 & 22.7 & 19.6 & 19.6\\
    & CLIP-$\mathcal{C}$ & \bf{15.7} & \bf{58.0} &  \bf{28.5} & \bf{50.1} &  11.4 & \bf{14.2}   & 0.7 & \bf{27.8} & \bf{86.8}  & \bf{26.1} & \bf{21.3} & \bf{21.2} \\
    \midrule
    \multirow{2}{*}{CC12M} &CLIP & 46.9 & \bf{78.0}  &  43.0 &  \bf{76.2} & 57.2 &  19.3 & 4.8 & \bf{41.2} & 89.7 & 33.8  & 27.8 & 37.9 \\
    & CLIP-$\mathcal{C}$ & \bf{48.1}  & 76.8 & \bf{44.8} & 73.5  & \bf{60.8} & \bf{21.9}  &  \bf{5.0} & 41.1 & \bf{90.3} & \bf{36.2} & \bf{36.1} & \bf{38.5}\\
    \midrule
      \multirow{2}{*}{RedCaps} & CLIP &78.8 &  72.8 & 38.7  & \bf{72.1} & 76.0 & 16.2 & 6.1 & 27.5 & 92.9 & 36.5 & 30.9 & 40.7 \\
       & CLIP-$\mathcal{C}$ &  \bf{79.0} & \bf{73.7} & \bf{42.2} & \bf{72.1} &  \bf{77.1} & \bf{18.1} & \bf{6.6} & \bf{29.4} & \bf{94.2} & \bf{41.1} & \bf{34.8}  & \bf{41.6} \\
    \bottomrule
     \end{tabular}
 \end{table*}
 
\subsection{Zero-shot Image Classification}
\label{sec: zero-shot-results}
We conduct a thorough study of the transfer learning capabilities of our model in zero-shot image classification on many downstream benchmarks, including ImageNet~\cite{imagenet} in \cref{tab:zero-shot-results-main}. Across different pretraining datasets, our method substantially improves over CLIP.  For ViT-S/16, CLIP-$\mathcal{C}$ achieves a $2\%$ top-1 improvement over the baseline CLIP model on ImageNet while outperforming CLIP on $12$ out of $12$ downstream datasets when pretraining on CC3M. Furthermore, these enhancements are maintained when we scale the vision encoder from ViT-S/16 to ViT-B/16 showing the continued effectiveness of our method over CLIP in a bigger model. When pretraining on RedCaps and CC12M, the gains of CLIP-$\mathcal{C}$ over CLIP on ImageNet are respectively are $0.9\%$ and $0.4\%$. These results are remarkable, considering that our approach and CLIP both use the same number of parameters, memory, and computational resources during pretraining. Even in the relatively data-rich settings of CC12M and RedCaps, CLIP-$\mathcal{C}$ still improves over CLIP on $11$ out of $12$  benchmarks for RedCaps and $9$ of the $12$ benchmarks for CC12M. 


\begin{table}[t]
    \caption{\textbf{Zero-shot Cross-modal Retrieval}. $\rho$ is set to $0.3$ for CC3M and $0.15$ for CC12M abd RedCaps. Similarly to zero-shot classification, our semantic composition model is nontrivially better than CLIP on zero-shot retrieval.}
     \label{tab:retrieval-results-main}
      \centering
     \begin{tabular}{l|l|cccc|cccc}
     & & \multicolumn{4}{c|}{Flickr30k} & \multicolumn{4}{c} {MS-COCO}\\
     \toprule
       \multirow{2}{*}{PT Dataset} & \multirow{2}{*}{Method} &  \multicolumn{2}{c}{Image $\rightarrow$ Text} & \multicolumn{2}{c|}{Text $\rightarrow$ Image}  & \multicolumn{2}{c}{Image $\rightarrow$ Text} & \multicolumn{2}{c}{Text $\rightarrow$ Image} \\
      & & R@1 & R@5 & R@1 & R@5  & R@1 & R@5  & R@1 & R@5  \\
     \midrule
      \multicolumn{10}{c}{\textit{Vision Encoder: ViT-S/16}} \\
      \midrule
     \multirow{2}{*}{CC3M} & CLIP & 35.2 &  62.3 & 25.4 & 49.12 & 17.3  & 39.0 & 13.1 & 31.2 \\
     &  CLIP-$\mathcal{C}$ & \bf{40.7}  &  \bf{70.9} & \bf{30.6}  & \bf{57.9} & \bf{21.4} & \bf{45.6}  & \bf{16.2} & \bf{36.5}\\
     \midrule
     \multicolumn{10}{c}{\textit{Vision Encoder: ViT-B/16}} \\
     \midrule
     \multirow{2}{*}{CC3M} & CLIP & 36.1 &  65.1 & 26.3  &  52.4 & 18.6&  41.1 & 13.9  & 32.8 \\
     &  CLIP-$\mathcal{C}$ &  \bf{39.6} & \bf{69.4}  &  \bf{31.2} & \bf{58.3} & \bf{22.9} & \bf{46.7} & \bf{17.0} & \bf{37.9} \\
     \midrule
      \multirow{2}{*}{CC12M} & CLIP & 61.5 & 87.2 & 46.1 & 74.9 & 36.2  &  64.2  &  25.3 & 49.7\\
  
    &  CLIP-$\mathcal{C}$ & \bf{66.0}  & \bf{87.8} & \bf{49.5} &  \bf{75.6} & \bf{38.4} & \bf{65.6} & \bf{26.4} & \bf{51.5} \\
     \midrule
      \multirow{2}{*}{RedCaps} & CLIP & 26.8 & 51.9 & 20.5 &  42.5 & 24.3  & 44.8 &  16.7 & 35.7 \\
     &  CLIP-$\mathcal{C}$ & \bf{32.3} &  \bf{57.2} &  \bf{23.6} & \bf{44.9} & \bf{27.1} & \bf{49.2} & \bf{18.2}  & \bf{38.4}  \\
    \bottomrule
     \end{tabular}
 \end{table}

\begin{table}[t]
     \caption{\textbf{Linear Probing}: CLIP-$\mathcal{C}$ (ours) is very competitive with CLIP in linear probe experiments on CC12M and RedCaps. Our method outperforms CLIP on a majority of benchmarks when using CC3M. Additionally, on our largest downstream dataset, ImageNet, CLIP-$\mathcal{C}$ beats CLIP in all settings except when pretraining on CC12M. All CLIP-$\mathcal{C}$ models here are trained using a sampling probability $\rho = 0.3$. }
     \label{tab:linear-probe-results-main}
     \centering
     \begin{tabular}{l|l|ccccccccccc|c}
     \rotatebox{90}{PT Dataset} & \rotatebox{90}{Method} & \rotatebox{90}{Food-101} & \rotatebox{90}{CIFAR-10} & \rotatebox{90}{CIFAR-100} & \rotatebox{90}{Caltech-101}  & \rotatebox{90}{Pets} &\rotatebox{90}{DTD}  & \rotatebox{90}{Country211} & \rotatebox{90}{Sun397}& \rotatebox{90}{STL-10} &  \rotatebox{90}{RESISC45} & \rotatebox{90}{EuroSAT} & \rotatebox{90}{ImageNet}\\
     \toprule
     \multicolumn{14}{c}{\textit{Vision Encoder: ViT-S/16}} \\
    \midrule
    \multirow{2}{*}{CC3M} & CLIP &  63.7 &  84.9 &  65.7 & 79.6 & 69.6 & 60.9 & 12.3 & 63.5 & 91.6 & 88.5 &  95.8 & 55.0\\
    & CLIP-$\mathcal{C}$ & 64.7 & 85.3 &  66.6 & 81.2 & 69.3  &  61.8 & 12.7 & 64.6 & 92.9 & 88.1  & 95.5  & \bf{56.8}\\
    \midrule
    \multicolumn{14}{c}{\textit{Vision Encoder: ViT-B/16}} \\
     \midrule
     \multirow{2}{*}{CC3M} & CLIP & 66.6 & 85.7  & 67.1  & 79.0 & 71.9 & 59.1 & 12.6 & 63.9 & 91.8 & 89.4 & 96.3  & 58.4\\
    & CLIP-$\mathcal{C}$ & 67.6 & 86.2 &  67.1 & 81.0 & 72.4 &  62.2 & 13.7 &  66.1 & 93.2  & 90.0 &  95.9 & \bf{59.5} \\
    \midrule
    \multirow{2}{*}{CC12M} &CLIP & 79.4 &  91.7  &  74.9 & 88.8  & 83.4 & 67.4  & 16.6  & 72.3 & 95.2 &  91.6 & 97.0 & \bf{68.6}\\
    & CLIP-$\mathcal{C}$ &  79.7 & 91.2  &  75.0 &  88.7 &  83.3&  68.6  &  17.0 & 72.8 &   95.7 &   91.7 &  96.7 &  68.3 \\
    \midrule
    \multirow{2}{*}{RedCaps} & CLIP & 88.7 &  91.4 &  73.5 & 88.3  & 90.2 & 69.1 & 15.7 & 68.6 & 96.8 & 91.7 & 95.4 & 70.4\\
    & CLIP-$\mathcal{C}$ &  88.9 &  90.8 & 74.1  & 88.6  & 89.9  &  69.7  &  16.0 & 69.7 & 96.9 & 92.3  &  97.0 & \bf{70.7} \\
    \bottomrule
     \end{tabular}
 \end{table}
 
\subsection{Zero-shot Cross-Modal Retrieval}
\label{sec:retrieval}
In addition to the zero-shot transfer results as detailed in \cref{sec: zero-shot-results}, we also provide analysis of the performance of CLIP-$\mathcal{C}$ versus CLIP on zero-shot cross-modal retrieval in \cref{tab:retrieval-results-main}. For these evaluations, we use MS-COCO~\cite{ms-coco} and Flickr30k~\cite{flickr30} as the downstream benchmarks. As in the zero-shot transfer setting, CLIP-$\mathcal{C}$ yields significant improvements over the baseline model on both MS-COCO and Flickr30k across different pretraining datasets and model sizes. For example, when using CC3M as the pretraining dataset, our method outperforms CLIP by over $5\%$ absolute top-1 retrieval accuracy in both image-to-text and text-to-image retrieval. The enhancement on MS-COCO is $4\%$ on image-to-text and $3\%$ on text-to-image retrievals. For both CLIP and our method, we noticed low retrieval results when pretraining on RedCaps, which we believe is related to the data distribution. We leave that analysis out for later works. 
 
\subsection{Linear Probe Evaluations}
\label{sec:linear-probe}

 Having verified the efficacy of our method using joint-embedding features in the zero-shot settings, we conduct several linear-probing evaluations in \cref{tab:linear-probe-results-main} to test the quality of the learned image features.  In these experiments, a randomly initialized linear layer is added on top of the pretrained image encoder $f_I$ which is frozen. The text encoder $f_T$, along with linear projections $g_I$ and $g_T$ are discarded. We train the linear layer for 50 epochs using a stochastic gradient descent optimizer with a weight decay of $0$ and a momentum of $0.9$. The linear probe learning rate and mini-batch size for each downstream dataset are provided in \cref{tab:linear-probe-hyper-parameters-sup} in the Supplemental.

 Using our proposed compositions, CLIP-$\mathcal{C}$ surpasses CLIP in several linear probe experiments when the pretraining dataset is relatively small, \eg, CC3M, indicating that our method also learns more discriminative image features than CLIP in that regime. These superior generalization capabilities come as a result of exposing the image encoder to a more diverse array of images through the compositions. We stress again that these gains are obtained using the same pretraining datasets (without any external augmentations), computational costs, and number of parameters as the CLIP baseline. The linear probing results are competitive on all downstream benchmarks when using larger pretraining datasets.

\section{Ablations}
\label{sec:ablations-main}
We ablate the various components of our framework including (1) providing more training resources to the CLIP model, (2) the sampling probability $\rho$, (3) semantic versus stylistic compositions, (4) the impact of stochasticity in drawing the second example, and (5) the composition function used for the images. We cover further ablations in \cref{sec:additional-ablations-sup} of the Supplemental detailing different mechanisms of pairing the examples, and other ways of merging captions including large language model rewrites. 

These ablation experiments underscore the importance of using semantically diverse examples in compositions (\cref{subsec:semantic-vrs-stylistic}). They also reveal that while incorporating a proportion of CLIP-$\mathcal{C}$ examples in the mini-batch contributes positively to performance, exclusively using such compositions during training detracts from downstream transfer capabilities (\cref{subsec:composition-ratio}). Finally, the results demonstrate the necessity of generating compound examples dynamically during training rather than relying on a static set of pre-generated instances (\cref{subsec:stochasticity-vrs-fixed-main}). Collectively, these insights affirm the effectiveness of the design principles underpinning our method.

Due to computational constraints, all ablation experiments are conducted using CC3M with ViT-S/16 as the image encoder. Additionally, we present only the zero-shot results of CIFAR-10, CIFAR-100, and ImageNet for the ablation experiments.  Also, we are unable to conduct multiple runs of each experiment because of the number and scale of our ablations.  As a result, we execute most of our experiments once using a shared fixed random seed.
\begin{table}[t]
     \caption{CLIP-$\mathcal{C}$ beats a CLIP despite using half the batch size of the CLIP model.}
     \label{tab:zero-shot-batch-size-ablation}
     \centering
     \begin{tabular}{l|c|ccc}
       Method & Batch-Size  &  CIFAR-10 & CIFAR-100 &  ImageNet \\
     \toprule
    CLIP & $2,048$ & 56.1 & 22.7 & 18.5 \\
   CLIP-$\mathcal{C}$ (Ours) & $1,024$ & \bf{67.7} &  \bf{31.1} & \bf{20.1} \\
    \bottomrule
     \end{tabular}
 \end{table}    
 
\setlength{\intextsep}{0pt}%
\begin{wraptable}{R}{0.55\textwidth}
    \caption{Both CLIP and CLIP-$\mathcal{C}$ are consistent across the three initializations. }
    \vspace{0.1cm}
     \label{tab:means-standard-deviation}
     \centering
     \begin{tabular}{l|ccc}
     Method  & CIFAR-10 & CIFAR-100 & ImageNet \\
     \toprule
    CLIP  &  $57.8\pm 2.36$ & $24.6\pm 1.62$ & $18.6\pm 0.24$\\
    CLIP-$\mathcal{C}$ & $64.7\pm 2.04$  & $27.6\pm 0.53$ & $20.4\pm 0.35$\\
    \bottomrule
     \end{tabular}
\end{wraptable}
 To ablate the impact of this choice, we train three models each for CLIP and CLIP-$\mathcal{C}$ on CC3M with three different random seeds. The results in~\cref{tab:means-standard-deviation} indicate that zero-shot performances are \textit{consistent} across different random initializations.
 
\subsection{Is CLIP-\texorpdfstring{$\mathcal{C}$}{$C$} a CLIP Model Exposed to More Data?}
\label{sec:resource-exposure}
\begin{figure}[t]
\begin{minipage}{0.47\textwidth}
    \centering
    \includegraphics[width=1\linewidth]{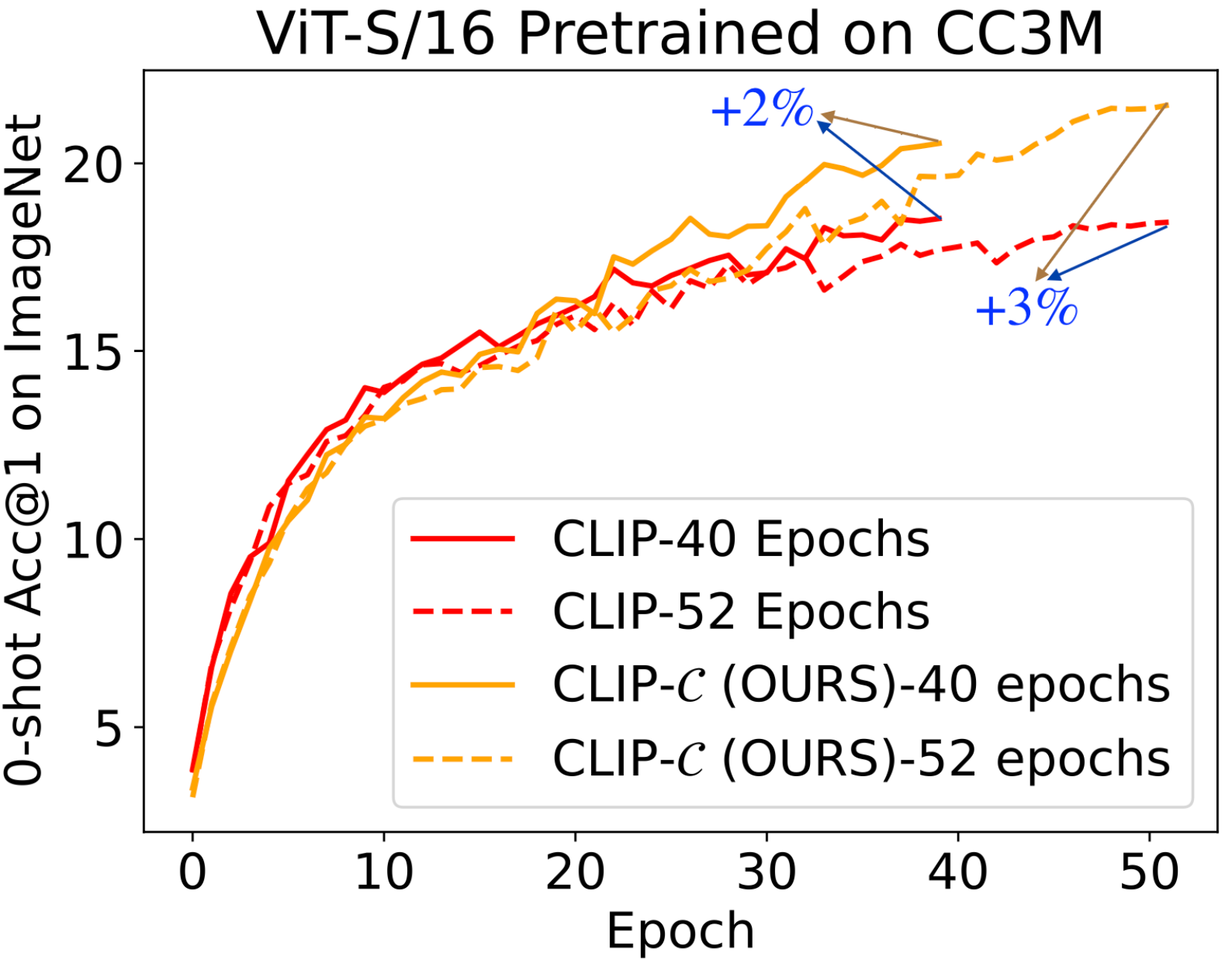}
   \caption{CLIP-$\mathcal{C}$ v.s. CLIP. Pretraining CLIP longer than CLIP-$\mathcal{C}$ does not close the performance gap. CLIP-$\mathcal{C}$ becomes even more superior as training duration increases.} 
   \label{fig:per-epoch-zero-shot-acc}
\end{minipage}
\hfill
\begin{minipage}{0.47\textwidth}
    \centering
    \includegraphics[width=1\textwidth]{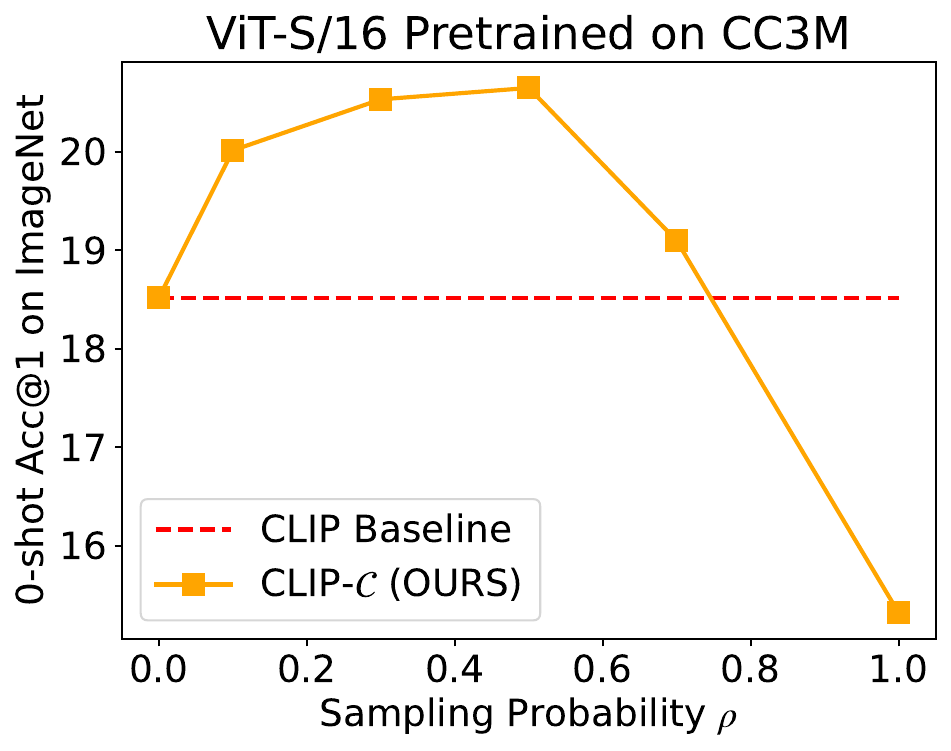}
    \caption{\textbf{Sampling probability $\rho$}. Our method is very effective when between 10\% and 50\% of the mini-batch are CLIP-$\mathcal{C}$ compositions but performs poorly when the entire batch is composite instances.}
    \label{fig:sampling-probability}
\end{minipage}
\end{figure}

 It could be argued that our method sees a lot more examples due to the compositions we employ, and that may be the reason for the observed improved performances. In \cref{tab:zero-shot-batch-size-ablation}, we show that a CLIP-$\mathcal{C}$ model that uses a batch size of $1,024$ examples outperforms the equivalent CLIP model trained with a batch-size of $2,048$ by 1.6\% on ImageNet,  strongly indicating that our method is different from ---and more impactful than--- a technique to increase the CLIP batch size. Similarly, we show in~\cref{fig:per-epoch-zero-shot-acc} that training CLIP for $(1 + \rho)$ times the number of epochs for the CLIP-$\mathcal{C}$ model does not close the performance gap (compare CLIP-52 and CLIP-$\mathcal{C}$-40 epochs in~\cref{fig:per-epoch-zero-shot-acc}). Indeed, the results in \cref{fig:per-epoch-zero-shot-acc} highlight the  \textbf{strong regularization effect} of CLIP-$\mathcal{C}$ as its superiority over CLIP emerges towards the later stages of pretraining. CLIP-$\mathcal{C}$ becomes even \textit{more superior to CLIP as training duration increases, extending the improvement from $+2\%$ zero-short accuracy on ImageNet when both models are trained for 40 epochs to over $+3\%$ when both are trained for 52 epochs.}  All these empirical results point to concrete beneficial qualities of CLIP-$\mathcal{C}$ as discussed in~\cref{sec:effectiveness} and not from any implicit amplified exposure to data.  

\subsection{Sampling Probability $\rho$}
\label{subsec:composition-ratio}
\vspace{-0.09cm}
The probability at which we create a composite sample as opposed to the original image-caption pair is an important parameter in our method which determines the percentage of the mini-batch that are compound instances. When $\rho = 0$, our method is identical to CLIP as no composition is performed. On the other extreme, when $\rho = 1$, all the examples in each mini-batch are instances of our composition method. As shown in \cref{fig:sampling-probability}, using a small non-zero sampling rate is more effective than CLIP. However, the performance deteriorates when more than fifty percent of the mini-batch are these compound image-text pairs. These results indicate that maintaining a reasonable percentage of the original examples is necessary likely because streamlined non-contradictory learning signal is significantly reduced when a majority of the batch are compositions. Also, since downstream evaluations do not involve such compositions, some exposure to examples with uniform semantic content during pretraining is important for effective transfer.  

\subsection{Why Semantic Compositions?}
\label{subsec:semantic-vrs-stylistic}

We call CLIP-$\mathcal{C}$ compositions semantic because the new instances are not just stylistically different from the constituent original examples, they are also semantically different.  Thus, it is fair to question whether or not this semantic differentiation is important in producing the observed favorable results over CLIP. After all, purely stylistic augmentations that use content from the same examples also increase data diversity and could yield the same outcomes as our semantic compositions. We investigate this prospect in this section. We train a model using two augmentations of the same example instead of two distinct examples as outlined in \cref{sec:method-ours}. On the image side, two random crops of the image are taken simulating two instances. For the text, we employ ``Easy Data Augmentation (EDA)''~\cite{wei-zou-2019-eda} to generate a caption for the second crop while the first crop uses the original caption. These two stylistically generated examples are then combined using CLIP-$\mathcal{C}$. 

\setlength{\intextsep}{0pt}%
\begin{wraptable}{R}{0.65\textwidth}
     \caption{\textbf{Semantic Compositions}: Using semantically distinct examples is better than stylistic augmentations. }
     \label{tab:zero-shot-stylistic}
     \vspace{0.2cm}
     \centering
     \begin{tabular}{l|ccc}
     Method  & CIFAR-10 & CIFAR-100 & ImageNet \\
     \toprule
     CLIP  & 56.1 &  22.7 & 18.5\\
    Stylistic &  53.7 & 25.0 & 19.0\\
    Semantic (Ours) & \bf{66.4}  & \bf{26.9} & \bf{20.5}\\
    \bottomrule
     \end{tabular}
     \vspace{0.2cm}
\end{wraptable}

In \cref{tab:zero-shot-stylistic}, it is evident that such stylistic augmentations are sub-optimal compared to the semantic generations we employ in CLIP-$\mathcal{C}$.  On ImageNet, the CLIP-$\mathcal{C}$ model achieves a $1.5\%$ absolute top-1 accuracy than the stylistic augmentations model. This suggests that the content of the new instances is important as the model prefers the use of distinct examples in the composition. We also note that just increasing the diversity of examples is helpful as the stylistic augmentations method yields a $0.5$\% zero-shot accuracy gain over CLIP on ImageNet.

\subsection{Impact of Stochasticity During Sampling}
\label{subsec:stochasticity-vrs-fixed-main}
Whenever CLIP-$\mathcal{C}$ composition is activated, the second example is usually chosen randomly from the dataset. This allows for every image-caption pair to be paired with any other image-caption pair in the dataset. Moreover, the pairings differ from one epoch to another, thus uncovering novel combinations of examples throughout pretraining.

\begin{wraptable}{R}{0.65\textwidth}
    \caption{\textbf{Impact of Stochasticity: }  Dynamic assignments is more effective than fixed pairings. }
     \label{tab:zero-shot-stochasticity}
     \vspace{0.2cm}
     \centering
     \begin{tabular}{l|ccc}
     Method  & CIFAR-10 & CIFAR-100 & ImageNet  \\
     \toprule
    CLIP  & 56.1 &  22.7 & 18.5\\
    Fixed &  60.4 & 28.0 & 19.7\\
    Dynamic (Ours) & \bf{66.4}  & \bf{26.9} & \bf{20.5}\\
    \bottomrule
     \end{tabular}
     \vspace{0.1cm}
\end{wraptable} 

We examine the impact of this dynamic nature of CLIP-$\mathcal{C}$ versus using fixed pairs of examples. To do this, for every example $x$, we allocate only one other example $x^\prime$ that is fixed throughout training. Then, whenever $x$ is involved in a CLIP-$\mathcal{C}$ composition, $x^\prime$ is used. The results in \cref{tab:zero-shot-stochasticity} suggest that dynamic compositions lead to better downstream results than fixed compositions. This makes intuitive sense because in the dynamic case, if a particular composition is unhelpful, there is a possibility of changing it in subsequent epochs. This possibility does not exist when the combinations are fixed.  in \cref{sec:sentece-embedding-sup} of the Supplemental, we also investigate scenarios whereby we combine examples whose captions are either close or far apart in the feature space of a sentence embedding model~\cite{wang2020minilm}.  


\subsection{Image Composition Function}
\label{sec:image-composition-main}
In this section, we compare our image mixing method with established systems such as CutMix~\cite{yun2019cutmix} and MixUP~\cite{zhang2018mixup}. When activated, MixUP executes a weighted pixel-wise summation of the two images, $\hat{x}_I^{(i)} = \omega \cdot x_I^{(i)} + (1 -\omega)\cdot x_I^{(j)}$ with the weighting factor, $\omega$ sampled from the beta distribution $\omega \sim \beta(1, 1)$. CutMix on the other hand takes a random crop from one of the images and pastes it at the same spatial location on the other image. The crop's dimensions are scaled by the value $\alpha = \sqrt{1 - \omega}$, $\omega \sim \beta(1, 1)$. That is, $H_{\text{cut}} = \alpha \cdot H$, $W_{\text{cut}} = \alpha \cdot W$ where $H$ and $W$ are the height and width of the image respectively. 
\setlength{\intextsep}{0pt}%
\begin{wraptable}{R}{0.65\textwidth}
     \caption{\textbf{Image Composition}: Our strategy outperforms CutMix and MixUP.  }
     \label{tab:zero-shot-image-mixup}
     \vspace{0.3cm}
     \centering
     \begin{tabular}{l|ccc}
      Function  & CIFAR-10 &  CIFAR-100 &  ImageNet \\
     \toprule
    MixUP~\cite{zhang2018mixup}  & 50.8  & 22.3 & 20.2\\
    CutMix~\cite{yun2019cutmix} & 54.2 &  \bf{26.9} & 20.4\\
    CLIP-$\mathcal{C}$ (Ours) & \bf{66.4}  & \bf{26.9} & \bf{20.5}\\
    \bottomrule
    \end{tabular}
    \vspace{0.1cm}
 \end{wraptable}

 Unlike MixUP, our method as depicted in \cref{fig:main-method} preserves the integrity of each crop, and does not paste parts of one image on the other as in CutMix. Additionally, using the center-half crop of each image guarantees that substantial portions of both images are represented in the output image. We believe these characteristics of our method are important as demonstrated by its superior zero-shot results over MixUP and CutMix in \cref{tab:zero-shot-image-mixup}.

\section{Conclusion}
 We have demonstrated in this study that fast and simple compositions of different image-caption pairs can significantly enhance the effectiveness of language-supervised visual representation learning models. This approach proves particularly beneficial when pretraining on smaller datasets. Our comprehensive analysis shows that CLIP-$\mathcal{C}$, our proposed model, delivers substantial improvements in zero-shot learning tasks over the baseline CLIP model and performs robustly in linear evaluation settings. Our ablation studies provide crucial insights, emphasizing that the observed performance improvements stem not from a mere increase in data augmentation but from the strategic use of semantically distinct examples in compositions.  We anticipate that these findings will encourage further exploration into novel and efficient uses of small-scale datasets for vision-language pretraining, especially in settings where it is difficult to curate massive amounts of paired data (\eg, medical and satellite images).


 
 
%
%
\bibliographystyle{splncs04}
\bibliography{main}

\begin{thebibliography}{10}
\providecommand{\url}[1]{\texttt{#1}}
\providecommand{\urlprefix}{URL }
\providecommand{\doi}[1]{https://doi.org/#1}

\bibitem{gpt3}
Brown, T., Mann, B., Ryder, N., Subbiah, M., Kaplan, J.D., Dhariwal, P.,
  Neelakantan, A., Shyam, P., Sastry, G., Askell, A., Agarwal, S.,
  Herbert-Voss, A., Krueger, G., Henighan, T., Child, R., Ramesh, A., Ziegler,
  D., Wu, J., Winter, C., Hesse, C., Chen, M., Sigler, E., Litwin, M., Gray,
  S., Chess, B., Clark, J., Berner, C., McCandlish, S., Radford, A., Sutskever,
  I., Amodei, D.: Language models are few-shot learners. In: Larochelle, H.,
  Ranzato, M., Hadsell, R., Balcan, M., Lin, H. (eds.) Advances in Neural
  Information Processing Systems. vol.~33, pp. 1877--1901 (2020),
  \url{https://proceedings.neurips.cc/paper_files/paper/2020/file/1457c0d6bfcb4967418bfb8ac142f64a-Paper.pdf}

\bibitem{multiview}
Caron, M., Misra, I., Mairal, J., Goyal, P., Bojanowski, P., Joulin, A.:
  Unsupervised learning of visual features by contrasting cluster assignments.
  In: Larochelle, H., Ranzato, M., Hadsell, R., Balcan, M., Lin, H. (eds.)
  Advances in Neural Information Processing Systems. vol.~33, pp. 9912--9924
  (2020),
  \url{https://proceedings.neurips.cc/paper_files/paper/2020/file/70feb62b69f16e0238f741fab228fec2-Paper.pdf}

\bibitem{cc12m}
Changpinyo, S., Sharma, P., Ding, N., Soricut, R.: {Conceptual 12M}: Pushing
  web-scale image-text pre-training to recognize long-tail visual concepts. In:
  CVPR (2021)

\bibitem{simclr}
Chen, T., Kornblith, S., Norouzi, M., Hinton, G.: A simple framework for
  contrastive learning of visual representations. In: III, H.D., Singh, A.
  (eds.) Proceedings of the 37th International Conference on Machine Learning.
  Proceedings of Machine Learning Research, vol.~119, pp. 1597--1607. PMLR
  (13--18 Jul 2020), \url{https://proceedings.mlr.press/v119/chen20j.html}

\bibitem{SimSiam2021}
Chen, X., He, K.: Exploring simple siamese representation learning. In: 2021
  IEEE/CVF Conference on Computer Vision and Pattern Recognition (CVPR). pp.
  15745--15753 (2021). \doi{10.1109/CVPR46437.2021.01549}

\bibitem{resisc45}
Cheng, G., Han, J., Lu, X.: Remote sensing image scene classification:
  Benchmark and state of the art. Proceedings of the IEEE  \textbf{105}(10),
  1865--1883 (Oct 2017). \doi{10.1109/jproc.2017.2675998},
  \url{http://dx.doi.org/10.1109/JPROC.2017.2675998}

\bibitem{dtd}
Cimpoi, M., Maji, S., Kokkinos, I., Mohamed, S., Vedaldi, A.: Describing
  textures in the wild. In: Proceedings of the {IEEE} Conf. on Computer Vision
  and Pattern Recognition ({CVPR}) (2014)

\bibitem{coates2011stl10}
Coates, A., Ng, A., Lee, H.: {An Analysis of Single Layer Networks in
  Unsupervised Feature Learning}. In: AISTATS (2011)

\bibitem{imagenet}
Deng, J., Dong, W., Socher, R., Li, L.J., Li, K., Fei-Fei, L.: Imagenet: A
  large-scale hierarchical image database. In: 2009 IEEE Conference on Computer
  Vision and Pattern Recognition. pp. 248--255 (2009).
  \doi{10.1109/CVPR.2009.5206848}

\bibitem{desai2021redcaps}
Desai, K., Kaul, G., Aysola, Z.T., Johnson, J.: Redcaps: Web-curated image-text
  data created by the people, for the people. In: Thirty-fifth Conference on
  Neural Information Processing Systems Datasets and Benchmarks Track (Round 1)
  (2021), \url{https://openreview.net/forum?id=VjJxBi1p9zh}

\bibitem{bert}
Devlin, J., Chang, M.W., Lee, K., Toutanova, K.: {BERT}: Pre-training of deep
  bidirectional transformers for language understanding. In: Proceedings of the
  2019 Conference of the North {A}merican Chapter of the Association for
  Computational Linguistics: Human Language Technologies, Volume 1 (Long and
  Short Papers). pp. 4171--4186. Association for Computational Linguistics (Jun
  2019). \doi{10.18653/v1/N19-1423}, \url{https://aclanthology.org/N19-1423}

\bibitem{dosovitskiy2021an}
Dosovitskiy, A., Beyer, L., Kolesnikov, A., Weissenborn, D., Zhai, X.,
  Unterthiner, T., Dehghani, M., Minderer, M., Heigold, G., Gelly, S.,
  Uszkoreit, J., Houlsby, N.: An image is worth 16x16 words: Transformers for
  image recognition at scale. In: International Conference on Learning
  Representations (2021), \url{https://openreview.net/forum?id=YicbFdNTTy}

\bibitem{fan2023improving}
Fan, L., Krishnan, D., Isola, P., Katabi, D., Tian, Y.: Improving clip training
  with language rewrites. arXiv:2305.20088  (2023)

\bibitem{caltech101}
Fei-Fei, L., Fergus, R., Perona, P.: Learning generative visual models from few
  training examples: An incremental bayesian approach tested on 101 object
  categories. CVPR Workshop  (2004)

\bibitem{devise2013}
Frome, A., Corrado, G.S., Shlens, J., Bengio, S., Dean, J., Ranzato, M.A.,
  Mikolov, T.: Devise: A deep visual-semantic embedding model. In: Burges, C.,
  Bottou, L., Welling, M., Ghahramani, Z., Weinberger, K. (eds.) Advances in
  Neural Information Processing Systems. vol.~26 (2013),
  \url{https://proceedings.neurips.cc/paper_files/paper/2013/file/7cce53cf90577442771720a370c3c723-Paper.pdf}

\bibitem{moco}
He, K., Fan, H., Wu, Y., Xie, S., Girshick, R.: Momentum contrast for
  unsupervised visual representation learning. In: 2020 IEEE/CVF Conference on
  Computer Vision and Pattern Recognition (CVPR). pp. 9726--9735 (2020).
  \doi{10.1109/CVPR42600.2020.00975}

\bibitem{resnet}
He, K., Zhang, X., Ren, S., Sun, J.: Deep residual learning for image
  recognition. In: 2016 IEEE Conference on Computer Vision and Pattern
  Recognition (CVPR). pp. 770--778 (2016). \doi{10.1109/CVPR.2016.90}

\bibitem{helber2017eurosat}
Helber, P., Bischke, B., Dengel, A., Borth, D.: Eurosat: A novel dataset and
  deep learning benchmark for land use and land cover classification (2017)

\bibitem{ITMix}
Hong, T., Guo, X., Ma, J.: Itmix: Image-text mix augmentation for transferring
  clip to image classification. In: 2022 16th IEEE International Conference on
  Signal Processing (ICSP). vol.~1, pp. 129--133 (2022).
  \doi{10.1109/ICSP56322.2022.9965292}

\bibitem{xu2023metaclip}
Hu~Xu, S.X., Tan, X.E., Huang, P.Y., Howes, R., Sharma, V., Li, S.W., Ghosh,
  G., Zettlemoyer, L., Feichtenhofer, C.: Demystifying clip data (2023)

\bibitem{openclip}
Ilharco, G., Wortsman, M., Wightman, R., Gordon, C., Carlini, N., Taori, R.,
  Dave, A., Shankar, V., Namkoong, H., Miller, J., Hajishirzi, H., Farhadi, A.,
  Schmidt, L.: Openclip (Jul 2021). \doi{10.5281/zenodo.5143773},
  \url{https://doi.org/10.5281/zenodo.5143773}

\bibitem{JiaAlign2021}
Jia, C., Yang, Y., Xia, Y., Chen, Y., Parekh, Z., Pham, H., Le, Q.V., Sung, Y.,
  Li, Z., Duerig, T.: Scaling up visual and vision-language representation
  learning with noisy text supervision. In: Meila, M., Zhang, T. (eds.)
  Proceedings of the 38th International Conference on Machine Learning, {ICML}
  2021. Proceedings of Machine Learning Research, vol.~139, pp. 4904--4916
  (2021), \url{http://proceedings.mlr.press/v139/jia21b.html}

\bibitem{jiang2023comclip}
Jiang, K., He, X., Xu, R., Wang, X.E.: Comclip: Training-free compositional
  image and text matching (2023)

\bibitem{Joulin2016}
Joulin, A., van~der Maaten, L., Jabri, A., Vasilache, N.: Learning visual
  features from large weakly supervised data. In: Leibe, B., Matas, J., Sebe,
  N., Welling, M. (eds.) Computer Vision -- ECCV 2016. pp. 67--84. Springer
  International Publishing (2016)

\bibitem{cifar}
Krizhevsky, A.: Learning multiple layers of features from tiny images. Tech.
  rep., University of Toronto (2009)

\bibitem{alexnet}
Krizhevsky, A., Sutskever, I., Hinton, G.E.: Imagenet classification with deep
  convolutional neural networks. In: Pereira, F., Burges, C., Bottou, L.,
  Weinberger, K. (eds.) Advances in Neural Information Processing Systems.
  vol.~25 (2012),
  \url{https://proceedings.neurips.cc/paper_files/paper/2012/file/c399862d3b9d6b76c8436e924a68c45b-Paper.pdf}

\bibitem{MaMMUT}
Kuo, W., Piergiovanni, A., Kim, D., Luo, X., Caine, B., Li, W., Ogale, A.,
  Zhou, L., Dai, A., Chen, Z., Cui, C., Angelova, A.: Mammut: A simple
  vision-encoder text-decoder architecture for multimodal tasks. Transactions
  on Machine Learning Research  (2023), \url{https://arxiv.org/abs/2303.16839}

\bibitem{lai2023scarcity}
Lai, Z., Zhang, H., Wu, W., Bai, H., Timofeev, A., Du, X., Gan, Z., Shan, J.,
  Chuah, C.N., Yang, Y., Cao, M.: From scarcity to efficiency: Improving clip
  training via visual-enriched captions (2023)

\bibitem{mindgap}
Lazaridou, A., Kuncoro, A., Gribovskaya, E., Agrawal, D., Liska, A., Terzi, T.,
  Gimenez, M., de~Masson~d\textquotesingle Autume, C., Kocisky, T., Ruder, S.,
  Yogatama, D., Cao, K., Young, S., Blunsom, P.: Mind the gap: Assessing
  temporal generalization in neural language models. In: Ranzato, M.,
  Beygelzimer, A., Dauphin, Y., Liang, P., Vaughan, J.W. (eds.) Advances in
  Neural Information Processing Systems. vol.~34, pp. 29348--29363 (2021),
  \url{https://proceedings.neurips.cc/paper_files/paper/2021/file/f5bf0ba0a17ef18f9607774722f5698c-Paper.pdf}

\bibitem{Li2017}
Li, A., Jabri, A., Joulin, A., van~der Maaten, L.: Learning visual n-grams from
  web data. In: 2017 IEEE International Conference on Computer Vision (ICCV).
  pp. 4193--4202. IEEE Computer Society (2017). \doi{10.1109/ICCV.2017.449},
  \url{https://doi.ieeecomputersociety.org/10.1109/ICCV.2017.449}

\bibitem{li2023blip2}
Li, J., Li, D., Savarese, S., Hoi, S.: {BLIP-2:} bootstrapping language-image
  pre-training with frozen image encoders and large language models. In: ICML
  (2023)

\bibitem{declip}
Li, Y., Liang, F., Zhao, L., Cui, Y., Ouyang, W., Shao, J., Yu, F., Yan, J.:
  Supervision exists everywhere: A data efficient contrastive language-image
  pre-training paradigm. In: International Conference on Learning
  Representations (2022), \url{https://openreview.net/forum?id=zq1iJkNk3uN}

\bibitem{ms-coco}
Lin, T.Y., Maire, M., Belongie, S., Hays, J., Perona, P., Ramanan, D.,
  Doll{\'a}r, P., Zitnick, C.L.: Microsoft coco: Common objects in context. In:
  Fleet, D., Pajdla, T., Schiele, B., Tuytelaars, T. (eds.) Computer Vision --
  ECCV 2014. pp. 740--755 (2014)

\bibitem{adamw}
Loshchilov, I., Hutter, F.: Decoupled weight decay regularization. In: ICML
  (2019), \url{https://openreview.net/forum?id=Bkg6RiCqY7}

\bibitem{mu2021slip}
Mu, N., Kirillov, A., Wagner, D., Xie, S.: Slip: Self-supervision meets
  language-image pre-training. arXiv:2112.12750  (2021)

\bibitem{naeem2023silc}
Naeem, M.F., Xian, Y., Zhai, X., Hoyer, L., Gool, L.V., Tombari, F.: Silc:
  Improving vision language pretraining with self-distillation (2023)

\bibitem{oord2019representation}
van~den Oord, A., Li, Y., Vinyals, O.: Representation learning with contrastive
  predictive coding (2019)

\bibitem{oxfordpets}
Parkhi, O.M., Vedaldi, A., Zisserman, A., Jawahar, C.V.: Cats and dogs. In:
  IEEE Conference on Computer Vision and Pattern Recognition (2012)

\bibitem{pytorch}
Paszke, A., Gross, S., Massa, F., Lerer, A., Bradbury, J., Chanan, G., Killeen,
  T., Lin, Z., Gimelshein, N., Antiga, L., Desmaison, A., Kopf, A., Yang, E.,
  DeVito, Z., Raison, M., Tejani, A., Chilamkurthy, S., Steiner, B., Fang, L.,
  Bai, J., Chintala, S.: Pytorch: An imperative style, high-performance deep
  learning library. In: Advances in Neural Information Processing Systems 32,
  pp. 8024--8035 (2019),
  \url{http://papers.neurips.cc/paper/9015-pytorch-an-imperative-style-high-performance-deep-learning-library.pdf}

\bibitem{Quattoni2007}
Quattoni, A., Collins, M., Darrell, T.: Learning visual representations using
  images with captions. In: 2007 IEEE Conference on Computer Vision and Pattern
  Recognition. pp.~1--8 (2007). \doi{10.1109/CVPR.2007.383173}

\bibitem{radford2021learning}
Radford, A., Kim, J.W., Hallacy, C., Ramesh, A., Goh, G., Agarwal, S., Sastry,
  G., Askell, A., Mishkin, P., Clark, J., Krueger, G., Sutskever, I.: Learning
  transferable visual models from natural language supervision (2021)

\bibitem{gpt1}
Radford, A., Narasimhan, K.: Improving language understanding by generative
  pre-training (2018)

\bibitem{radford2019language}
Radford, A., Wu, J., Child, R., Luan, D., Amodei, D., Sutskever, I.: Language
  models are unsupervised multitask learners  (2019)

\bibitem{cc3m}
Sharma, P., Ding, N., Goodman, S., Soricut, R.: Conceptual captions: A cleaned,
  hypernymed, image alt-text dataset for automatic image captioning. In:
  Proceedings of the 56th Annual Meeting of the Association for Computational
  Linguistics (Volume 1: Long Papers). pp. 2556--2565. Association for
  Computational Linguistics (Jul 2018). \doi{10.18653/v1/P18-1238},
  \url{https://aclanthology.org/P18-1238}

\bibitem{deit}
Touvron, H., Cord, M., Douze, M., Massa, F., Sablayrolles, A., Jegou, H.:
  Training data-efficient image transformers \&; distillation through
  attention. In: Meila, M., Zhang, T. (eds.) Proceedings of the 38th
  International Conference on Machine Learning. Proceedings of Machine Learning
  Research, vol.~139, pp. 10347--10357. PMLR (18--24 Jul 2021),
  \url{https://proceedings.mlr.press/v139/touvron21a.html}

\bibitem{touvron2023llama}
Touvron, H., Martin, L., Stone, K., Albert, P., Almahairi, A., Babaei, Y.,
  Bashlykov, N., Batra, S., Bhargava, P., Bhosale, S., Bikel, D., Blecher, L.,
  Ferrer, C.C., Chen, M., Cucurull, G., Esiobu, D., Fernandes, J., Fu, J., Fu,
  W., Fuller, B., Gao, C., Goswami, V., Goyal, N., Hartshorn, A., Hosseini, S.,
  Hou, R., Inan, H., Kardas, M., Kerkez, V., Khabsa, M., Kloumann, I., Korenev,
  A., Koura, P.S., Lachaux, M.A., Lavril, T., Lee, J., Liskovich, D., Lu, Y.,
  Mao, Y., Martinet, X., Mihaylov, T., Mishra, P., Molybog, I., Nie, Y.,
  Poulton, A., Reizenstein, J., Rungta, R., Saladi, K., Schelten, A., Silva,
  R., Smith, E.M., Subramanian, R., Tan, X.E., Tang, B., Taylor, R., Williams,
  A., Kuan, J.X., Xu, P., Yan, Z., Zarov, I., Zhang, Y., Fan, A., Kambadur, M.,
  Narang, S., Rodriguez, A., Stojnic, R., Edunov, S., Scialom, T.: Llama 2:
  Open foundation and fine-tuned chat models (2023)

\bibitem{wang2020minilm}
Wang, W., Wei, F., Dong, L., Bao, H., Yang, N., Zhou, M.: Minilm: Deep
  self-attention distillation for task-agnostic compression of pre-trained
  transformers (2020)

\bibitem{wei-zou-2019-eda}
Wei, J., Zou, K.: {EDA}: Easy data augmentation techniques for boosting
  performance on text classification tasks. In: Proceedings of the 2019
  Conference on Empirical Methods in Natural Language Processing and the 9th
  International Joint Conference on Natural Language Processing (EMNLP-IJCNLP).
  pp. 6382--6388. Association for Computational Linguistics (Nov 2019).
  \doi{10.18653/v1/D19-1670}, \url{https://aclanthology.org/D19-1670}

\bibitem{wolf-etal-2020-transformers}
Wolf, T., Debut, L., Sanh, V., Chaumond, J., Delangue, C., Moi, A., Cistac, P.,
  Rault, T., Louf, R., Funtowicz, M., Davison, J., Shleifer, S., von Platen,
  P., Ma, C., Jernite, Y., Plu, J., Xu, C., Le~Scao, T., Gugger, S., Drame, M.,
  Lhoest, Q., Rush, A.: Transformers: State-of-the-art natural language
  processing. In: Liu, Q., Schlangen, D. (eds.) Proceedings of the 2020
  Conference on Empirical Methods in Natural Language Processing: System
  Demonstrations. pp. 38--45. Association for Computational Linguistics (Oct
  2020). \doi{10.18653/v1/2020.emnlp-demos.6},
  \url{https://aclanthology.org/2020.emnlp-demos.6}

\bibitem{Otter}
Wu, B., Cheng, R., Zhang, P., Gao, T., Gonzalez, J.E., Vajda, P.: Data
  efficient language-supervised zero-shot recognition with optimal transport
  distillation. In: International Conference on Learning Representations
  (2022), \url{https://openreview.net/forum?id=G89-1yZLFHk}

\bibitem{sun397}
{Xiao}, J., {Hays}, J., {Ehinger}, K.A., {Oliva}, A., {Torralba}, A.: Sun
  database: Large-scale scene recognition from abbey to zoo. In: 2010 IEEE
  Computer Society Conference on Computer Vision and Pattern Recognition. pp.
  3485--3492 (June 2010). \doi{10.1109/CVPR.2010.5539970}

\bibitem{xu-etal-2021-videoclip}
Xu, H., Ghosh, G., Huang, P.Y., Okhonko, D., Aghajanyan, A., Metze, F.,
  Zettlemoyer, L., Feichtenhofer, C.: {V}ideo{CLIP}: Contrastive pre-training
  for zero-shot video-text understanding. In: Proceedings of the 2021
  Conference on Empirical Methods in Natural Language Processing. pp.
  6787--6800. Association for Computational Linguistics (Nov 2021).
  \doi{10.18653/v1/2021.emnlp-main.544},
  \url{https://aclanthology.org/2021.emnlp-main.544}

\bibitem{flickr30}
Young, P., Lai, A., Hodosh, M., Hockenmaier, J.: From image descriptions to
  visual denotations: New similarity metrics for semantic inference over event
  descriptions. Transactions of the Association for Computational Linguistics
  \textbf{2},  67--78 (2014). \doi{10.1162/tacl_a_00166},
  \url{https://aclanthology.org/Q14-1006}

\bibitem{yu2022coca}
Yu, J., Wang, Z., Vasudevan, V., Yeung, L., Seyedhosseini, M., Wu, Y.: Coca:
  Contrastive captioners are image-text foundation models. Transactions on
  Machine Learning Research  (2022),
  \url{https://openreview.net/forum?id=Ee277P3AYC}

\bibitem{yuksekgonul2023when}
Yuksekgonul, M., Bianchi, F., Kalluri, P., Jurafsky, D., Zou, J.: When and why
  vision-language models behave like bags-of-words, and what to do about it?
  In: The Eleventh International Conference on Learning Representations (2023),
  \url{https://openreview.net/forum?id=KRLUvxh8uaX}

\bibitem{yun2019cutmix}
Yun, S., Han, D., Oh, S.J., Chun, S., Choe, J., Yoo, Y.: Cutmix: Regularization
  strategy to train strong classifiers with localizable features. In: ICCV
  (2019)

\bibitem{lit2022}
Zhai, X., Wang, X., Mustafa, B., Steiner, A., Keysers, D., Kolesnikov, A.,
  Beyer, L.: Lit: Zero-shot transfer with locked-image text tuning. In:
  Proceedings of the IEEE/CVF Conference on Computer Vision and Pattern
  Recognition (CVPR). pp. 18123--18133 (June 2022)

\bibitem{zhang2018mixup}
Zhang, H., Cisse, M., Dauphin, Y.N., Lopez-Paz, D.: mixup: Beyond empirical
  risk minimization. In: International Conference on Learning Representations
  (2018), \url{https://openreview.net/forum?id=r1Ddp1-Rb}

\bibitem{zhao2022lavila}
Zhao, Y., Misra, I., Kr{\"a}henb{\"u}hl, P., Girdhar, R.: Learning video
  representations from large language models. In: arXiv:2212.04501 (2022)

\end{thebibliography}

\clearpage

\setcounter{page}{1}
\setcounter{section}{0}
\maketitlesupplementary
\renewcommand\thesection{\Alph{section}}

\begin{table*}
    \caption{\textbf{Linear probe evaluation hyper-parameters}: We use stochastic gradient descent optimizer with a decay of 0 and a momentum of 0.9 for all linear probe experiments. All linear probe experiments are tuned for 50 epochs. }
     \label{tab:linear-probe-hyper-parameters-sup}
     \centering
     \begin{tabular}{l|ccccccccccc|c}
     Hyper-parameter & \rotatebox{90}{Food-101} & \rotatebox{90}{CIFAR-10} & \rotatebox{90}{CIFAR-100} & \rotatebox{90}{Caltech-101} &  \rotatebox{90}{Pets} &\rotatebox{90}{DTD}  & \rotatebox{90}{Country211} & \rotatebox{90}{Sun397}& \rotatebox{90}{STL-10} &  \rotatebox{90}{RESISC45} & \rotatebox{90}{EuroSAT} & \rotatebox{90}{ImageNet}\\
     \toprule
    Batch Size & 16  &  16 &  64 &  64 & 16 & 16 & 64 & 16 & 16 & 16 & 16 & 1024 \\
    Learning Rate & 0.1 &  0.1  &  0.1 & 0.05   &  0.1 & 0.1  & 0.05 & 0.05 &  0.1 & 0.1 &  0.1&  0.1 \\
    \bottomrule
     \end{tabular}
 \end{table*}
\section{Implementation Details}
\label{sec:hyper-parameters-sup}
We discuss other hyper-parameters in our setup in addition to those described in \cref{sec: experimental-setup-main}. During pretraining, AdamW~\cite{adamw} with $\beta_1 = 0.9$ and $\beta_2 = 0.98$ is used as the optimizer. We use a Stochastic Gradient Descent optimizer with a momentum of 0.9 for linear probing. The learning rate schedule is always set to Cosine Annealing with a warm-up period of $5$ epochs. In all pretraining settings, we train for 40 epochs and use the best checkpoint determined by the top-1 zero-shot accuracy on ImageNet~\cite{imagenet} for downstream evaluations. In linear probing experiments, we use the final results after training for 50 epochs on the respective dataset.  

In pretraining, the batch-size and learning rate are set to $2,048$ and $0.003$, respectively. For linear probe experiments, we use a batch-size of 1024 and a learning rate of $0.1$ when evaluating on ImageNet. For all other downstream datasets, we select the batch-size and learning rate using the ViT-B/16 CC3M pretrained model as follows. First, we choose the best batch-size from the set $\{16, 64, 256\}$ before selecting a learning rate from the set $\{0.05, 0.1, 0.2, 0.3\}$ using the chosen batch-size.  After selecting a batch-size and learning rate using the CC3M ViT/B-16 model, we then apply those parameters to all other models without any further hyperparameter searches. \Cref{tab:linear-probe-hyper-parameters-sup} contains details about the combinations of batch-sizes and base learning rates we use for linear probing on all downstream datasets. Every linear experiment run is executed on a single NVIDIA RTX A6000 GPU. 

The image resolution is always set to $224 \times 224$. Given an image of arbitrary size, we invoke the ``RandomResizedCrop'' from PyTorch~\cite{pytorch} with bicubic interpolation to generate an appropriately sized crop for training. The ``scale'' argument of ``RandomResizedCrop'' is set to $(0.6, 1.0)$ during pretraining and $(0.08, 1.0)$ during linear probing.  In linear evaluations, the output crop is further flipped 
 horizontally with a probability of $0.5$. Following previous works, we normalize all images using the ImageNet mean and standard deviation values irrespective of the dataset. We do not use cropping in the zero-shot classification and retrieval settings --- we simply resize the image to $224 \times 224$ before normalization.  In all our experiments, we used a fixed seed for the PyTorch and Numpy random number generators. We also enabled PyTorch's CUDNN deterministic setting.

 \begin{table}[t]
    \caption{\textbf{Pretraining Datasets: } Size is the number of image-text pairs in the dataset. Min, Avg, and Max are respectively the minimum, average, and maximum number of words in a caption obtained by splitting the string at spaces. CC12M tends to have the longest captions on average.}
    \label{tab:pre-training}
    \centering
    \begin{tabular}{lcccc}
        \toprule
         Dataset & Size & Min. & Avg. & Max. \\
         \midrule
        CC3M~\cite{cc3m} & 2.85M & 4 & 10.25 & 50 \\
        CC12M~\cite{cc12m} & 9.55M & 1 & 17.76 & 242 \\
        RedCaps~\cite{desai2021redcaps} & 12.01M & 1 & 9.50 & 70\\
        \bottomrule
    \end{tabular}
\end{table}

 \begin{table}[t]
      \caption{\textbf{Downstream Datasets: } Train and Test respectively denote the sizes of the training and evaluation sets. ``Classes'' is the number of categories while ``Metric'' is the evaluation metric. ``Acc'' represents top-1 accuracy over the entire dataset and ``P/C Acc'' represents the mean of per-category top-1 accuracy.}
         \label{tab:downstream-datasets-sup}
     \centering
     \begin{tabular}{c|cccc}
     \toprule
        Dataset & Train & Test & Classes & Metric\\
        \midrule
         Food-101 & 75,750 & 25,250 & 101 & Acc\\
         CIFAR-10 & 50,000 & 10,000 & 10 & Acc\\
         CIFAR-100 & 50,000 & 10,000 & 100 & Acc\\
         Caltech-101 & 3,060 & 6,084 & 102 &  P/C Acc\\
         Pets & 3,721 & 3,669 & 37 &  P/C Acc\\
         DTD & 1,880 & 1,880 & 47 & Acc\\
         Country211 & 31,650 & 21,100 & 211 & Acc\\
         Sun397 & 76, 127 & 21,750 & 397 & Acc\\
         STL-10 & 5,000 & 8,000 & 10 & Acc\\
         RESISC45 & 25,200 & 6,300 & 45 & Acc\\
         EuroSAT & 21,600 & 5,400 & 10 & Acc \\
         ImageNet & 1,281,167 & 50,000 & 1000 & Acc\\
         \bottomrule
     \end{tabular}
 \end{table}

\section{Datasets}
\label{sec:datasets-sup}
We cover basic characteristics of our pretraining datasets in \cref{tab:pre-training} and the downstream benchmarks in \cref{tab:downstream-datasets-sup}. We downloaded the pretraining datasets images manually from the internet using the URLs provided in Sharma~\cite{cc3m}, Changpinyo \etal~\cite{cc12m}, and Desai \etal~\cite{resisc45}. As a result, we could not retrieve the full original datasets as some of the links are now broken. Most of the downstream datasets were downloaded from Tensorflow datasets\footnote{https://www.tensorflow.org/datasets/catalog/overview} and Torchvision datasets\footnote{https://pytorch.org/vision/stable/datasets.html}. 

 
\section{Additional Ablations}
\label{sec:additional-ablations-sup}

\begin{table}[t]
     \caption{\textbf{Modality Involved in Composition}: Applying semantic compositions on both modalities is the most consistently effective method across different downstream datasets and tasks. }
     \label{tab:zero-shot-modality-ablation}
     \centering
     \begin{tabular}{l|ccc}
     Modality  & CIFAR-10 & CIFAR-100 & ImageNet \\
     \toprule
    Text only  & 55.2 & 27.0  &21.3 \\
    Images only & 55.9 & 23.1 & 19.4\\
    Text \& Images & 66.4  & 26.9 & 20.5\\
    \bottomrule
     \end{tabular}
 \end{table}
 
\subsection{Impact of Modality Used in Composition}
\label{sec:modality-composition-sup}
Since our inputs are of different modalities, visual and textual, it is important to examine whether compositions in each of these modalities produce similar effects. To that end, in \cref{tab:zero-shot-modality-ablation}, we conduct analysis where our method is applied on (1) only the captions, (2) only the images, and (3) both the captions and images. Of these three variations, executing the compositions on both the captions and images is the most effective, probably due to the symmetry of transforming both modalities. The second most effective is the captions-only approach. Option (2) is the least effective method likely because the images are naturally augmented (random cropping) in the baseline method whereas the captions are fixed.  These observations suggest that our method is more helpful in learning representations of the texts relative to those of the images. They also help elucidate why we obtain much bigger improvements over the baseline in zero-shot settings compared to linear evaluations.

\begin{table*}[t]
    \caption{\textbf{Zero-shot Results on Ways of Pairing Examples}: For each row, we pretrain a ViT-S/16 model on the 100k CC3M subset explained in \cref{sec:sentece-embedding-sup}. \textbf{L-C-S} stands for Largest Caption-Similarity while \textbf{S-C-S} represents Smallest Caption-Similarity. \textbf{Random} is the case where we perform random assignments at the beginning and fix them for the rest of the training. We pretrain for 40 epochs using a reduced learning rate of $3\times 10^{-4}$, a batch-size of $256$, and a warm-up period to $20$ epochs. }
     \label{tab:zero-shot-other-pairings-sup}
     \centering
     \begin{tabular}{l|ccccccccccc|c}
   \rotatebox{90}{Method} & \rotatebox{90}{Food-101} & \rotatebox{90}{CIFAR-10} & \rotatebox{90}{CIFAR-100} & \rotatebox{90}{Caltech-101} & \rotatebox{90}{Pets} &\rotatebox{90}{DTD} & \rotatebox{90}{Country211} & \rotatebox{90}{Sun397}& \rotatebox{90}{STL-10} &  \rotatebox{90}{RESISC45} & \rotatebox{90}{EuroSAT} & \rotatebox{90}{ImageNet}\\
     \toprule
     CLIP &  1.55 &  18.1 &  4.93  &  8.73  & 2.64 & 2.98  & 0.36 & 0.6 & 28.7 & 4.97 & 20.4 & 1.43\\
     \midrule
    Random & 2.22 & 16.7 &  5.38 & 7.86  & 2.95 & 2.18 & 0.56 & 0.69 & 25.6  & 8.46 & 11.4 & 1.52 \\
    L-C-S & 2.63 & 20.4 &  5.66 & 9.36 & 2.64 & 2.71 & 0.38 &  1.12 & 27.5  &  8.31 & 19.2 & 1.56 \\
    S-C-S & 2.17 & 23.6 & 5.87  & 8.46 & 1.93 & 3.03 & 0.52 & 0.75 &  27.4 & 8.69 & 12.7 &  1.49\\
    \midrule
     CLIP-$\mathcal{C}$ & 1.83 & 21.9 &  5.68 & 8.15 & 2.74 & 3.19  & 0.37 & 0.84 & 25.3 & 4.90  & 12.3 & 1.61 \\
    \bottomrule
     \end{tabular}
 \end{table*}

\subsection{Other Ways of Pairing Examples}
\label{sec:sentece-embedding-sup}
In CLIP-$\mathcal{C}$, the second example is chosen randomly from the dataset whenever the composition is active. In this section, we explore other ways of choosing that second image-caption pair. The configurations studied here include selecting a second instance whose caption is (1) the closest to, or (2) the farthest from the first caption of the anchor, with distance measured by the pair-wise cosine similarity between features obtained from a sentence embedding model~\cite{wang2020minilm}. We also consider the scenario where examples are paired randomly. However, the pairings are fixed once generated at the beginning of training. (CLIP-$\mathcal{C}$ uses random assignments that change in every epoch.) 

Since computing the embeddings of all captions as well as the pair-wise cosine similarities of all examples is extraordinarily expensive computationally,  we randomly selected a subset of $100,000$ examples from CC3M for these studies. We then computed the normalized features for all captions using the All-MiniLM-L6-v2\footnote{https://huggingface.co/sentence-transformers/all-MiniLM-L6-v2} pretrained model from HuggingFace~\cite{wolf-etal-2020-transformers}. Afterward, we generated the $100k \times 100k$ pair-wise cosine similarities for this experiment. We also run CLIP, our method as discussed in the paper, and the fixed random assigned described above on this subset. 

In zero-shot evaluations in this setting, we do not observe any significant differences between pairing examples whose captions are close or far in the embedding space. More importantly, CLIP-$\mathcal{C}$ beats all other setups on a plurality of the downstream benchmarks including on ImageNet. This performance, along with its simplicity, efficiency, and scalability (since there is no need for pair-wise comparisons of examples) all further justify the adoption of dynamic random sampling in our method.

 \begin{table*}[t]
    \caption{\textbf{Zero-shot Results on Text Composition Methods}: For each row, we pretrain a ViT-S/16 model on the 100k CC3M subset detailed in \cref{sec: caption-composition-sup}. CLIP-$\mathcal{C}_{\text{w/ AND}}$ (default) represents the scenario where the captions are concatenated with ``and'' as a conjunction. CLIP-$\mathcal{C}_{\text{w/o AND}}$ denotes the case where we omit the ``and''. Finally, CLIP-$\mathcal{C}_{\text{LLM}}$ is the language model rewriting method.  We pretrain for 40 epochs using a reduced learning rate of $3\times 10^{-4}$, a batch-size of $256$, and a warm-up period to $20$ epochs. }
     \label{tab:text-composition-method}
     \centering
     \begin{tabular}{l|ccccccccccc|c}
   \rotatebox{90}{Method} & \rotatebox{90}{Food-101} & \rotatebox{90}{CIFAR-10} & \rotatebox{90}{CIFAR-100} & \rotatebox{90}{Caltech-101} & \rotatebox{90}{Pets} &\rotatebox{90}{DTD} & \rotatebox{90}{Country211} & \rotatebox{90}{Sun397}& \rotatebox{90}{STL-10} &  \rotatebox{90}{RESISC45} & \rotatebox{90}{EuroSAT} & \rotatebox{90}{ImageNet}\\
     \toprule
    CLIP-$\mathcal{C}_{\text{w/ AND}}$ & 1.92 &  15.6 & 5.58  & 8.49 & 4.35 & 3.24 & 0.50  & 1.15 & 31.0 & 8.97  & 19.8 & 1.70\\
    CLIP-$\mathcal{C}_{\text{w/o AND}}$ &  2.34 & 19.48 &  5.33 &  7.67 & 2.34 &  4.04  & 0.40 & 1.29 & 31.6 &  6.49 & 18.5 & 1.92 \\
    CLIP-$\mathcal{C}_{\text{LLM}}$  &  1.94 &  15.6 &  5.68 &  8.22   &  3.81 & 3.19  & 0.45 & 1.13 & 26.7 & 6.17 & 9.05 & 1.83\\
    \bottomrule
     \end{tabular}
 \end{table*}
\subsection{Different Methods of Merging Captions}
\label{sec: caption-composition-sup}
As in \cref{sec:image-composition-main} where we looked at different ways of composing the merged image, we study other ways of merging the captions in this section. In CLIP-$\mathcal{C}$, we adopt the very simple and highly flexible process of concatenating the two captions using ``and'' as a conjunction. Here, we compare that method against omitting the conjunction or rewriting the merged caption with a large language model (LLM). 

Similarly to the ablation in \cref{sec:sentece-embedding-sup}, we randomly selected $100, 000$ examples from CC3M for this ablation because of the costs involved in the LLM rewriting method. Each example in the subset is associated with a second example (drawn from the full dataset) at random. We generate these associations once and use them for all ablation methods in this section: when (1) concatenating \textit{with ``and''}, (2) concatenating \textit{without ``and''}, and (3) rewriting the output of (1) with an LLM. 

\begin{figure*}[t]
    \centering
    \includegraphics[width=\linewidth]{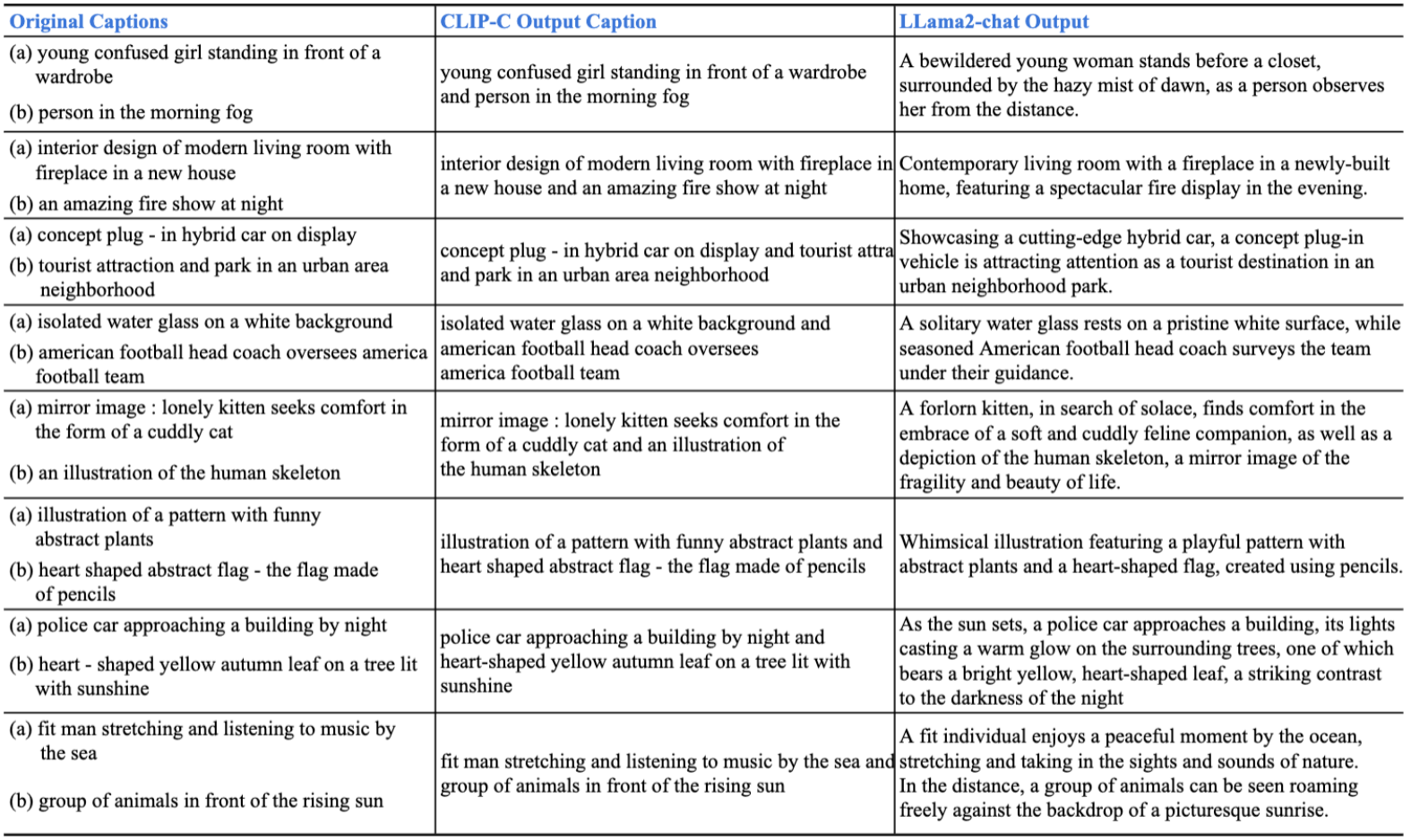}
    \caption{Sample of captions generated using the LLM-based method and our CLIP-$\mathcal{C}$ procedure.}
    \label{fig:example-captions}
\end{figure*}

We employ the open-source 70B parameter model LLama2-chat~\cite{touvron2023llama} for the LLM rewriting method. The temperature and top-p arguments of the model are set to $0.8$ and $0.95$, respectively. The maximum sequence length is set to 256. To rewrite a given caption, we (a) instruct the model to act as a text completion assistant, and (b) prompt the model as: \textit{rephrase the sentence \{merged caption\}} where merged caption is from option (1) above.  We use the first generated sentence as the new caption. The LLM-based captions are generated offline and saved to file before training begins. We emphasize that only $100,000$ rewrites are undertaken, one for each entry in the sampled subset. Some examples of these rewritten captions are provided in \cref{fig:example-captions}. 

On zero-shot classification in this limited setting, our simple concatenation method with and without ``and'' as a conjunction, each obtained the highest accuracy on $6$ out of $14$ downstream datasets while the LLM rewrite option achieved the highest performance on the remaining $2$ datasets as shown in \cref{tab:text-composition-method}. These results show that besides being highly efficient and flexible, our method is also comparable with the more resource-intensive large language model rewriting method. Additionally, while the LLM rewrite method may hallucinate non-grounded captions (See \cref{fig:example-captions}), our method is devoid of such hallucinations. We use concatenation with the conjunction as the default caption composition method because of its coherence.

\end{document}